\PassOptionsToPackage{authoryear,sort&compress,round}{natbib}
\documentclass[11pt,a4paper,logo,copyright]{googlecloud}

\usepackage{natbib}
\usepackage{nicefrac}
\usepackage{multicol}
\usepackage{multirow}
\usepackage{longtable}
\usepackage{array}
\usepackage{makecell}
\usepackage{bbding}
\usepackage{boldline}
\usepackage{wrapfig}
\usepackage{placeins}
\usepackage[export]{adjustbox}
\usepackage{fvextra}
\usepackage[normalem]{ulem}
\usepackage[most,breakable]{tcolorbox}
\usepackage{pgfplots}
\usepackage{arydshln}
\usepackage{subcaption}
\usepackage{siunitx}
\usepackage{titletoc}
\pgfplotsset{compat=1.18}
\usepgfplotslibrary{units}

\usepackage{amsmath,amsfonts,bm}

\def\eqref#1{equation~\ref{#1}}

\def\1{\bm{1}}

\DeclareMathAlphabet{\mathsfit}{\encodingdefault}{\sfdefault}{m}{sl}
\SetMathAlphabet{\mathsfit}{bold}{\encodingdefault}{\sfdefault}{bx}{n}

\usepackage{xspace}
\usepackage{fontawesome}

\definecolor{darkblue}{rgb}{0, 0, 0.5}
\definecolor{veronica-red}{RGB}{196,30,58}
\hypersetup{colorlinks=true, citecolor=darkblue, linkcolor=darkblue, urlcolor=darkblue}
\definecolor{ForestGreen}{RGB}{34,139,34}
\definecolor{BrickRed}{rgb}{.72,0,0}
\definecolor{LakeBlue}{RGB}{0,61,153}
\definecolor{RoyalBlue}{cmyk}{1,0.50,0,0}
\definecolor{boxgreen}{rgb}{0.9,1.0,0.9}
\definecolor{boxblue}{rgb}{0.9,0.9,1.0}
\definecolor{boxred}{rgb}{1.0, 0.85, 0.95}

\newcommand{\ours}{VeriCUA\xspace}

\usepackage{float}
\usepackage{needspace}
\let\caseoriginalsfdefault\sfdefault
\usepackage{tgheros}
\let\sfdefault\caseoriginalsfdefault

\definecolor{caseheadingblue}{HTML}{00205B}
\definecolor{casefill}{HTML}{F6F7F6}
\definecolor{casestroke}{HTML}{C7CECC}
\definecolor{casetext}{HTML}{2F3435}
\definecolor{caseaccent}{HTML}{566B7B}
\definecolor{casebluefill}{HTML}{E9EDF0}
\definecolor{casebluebar}{HTML}{71879A}
\definecolor{casewarmfill}{HTML}{F7F1EA}
\definecolor{casewarmstroke}{HTML}{DCCDBE}
\definecolor{casewarmaccent}{HTML}{957760}
\definecolor{casemet}{HTML}{2E7D4F}
\definecolor{caseunmet}{HTML}{B3443E}

\providecommand{\caseassetdir}{Figures/case_study}

\newcommand{\summarylabel}[1]{%
  {\fontsize{6.5}{7.6}\selectfont\color{caseaccent}\bfseries\MakeUppercase{#1}}}
\newcommand{\rmet}{\textcolor{casemet}{\ding{51}}}
\newcommand{\runmet}{\textcolor{caseunmet}{\ding{55}}}
\newcommand{\runknown}{\textcolor{caseaccent}{$\circ$}}
\newcommand{\rubricline}[2]{%
  \par\noindent\hangindent=1.15em\hangafter=1 #1\hspace{0.45em}#2}
\newcommand{\evidenceline}[1]{%
  \par\noindent\hangindent=1.15em\hangafter=1
  \textcolor{casewarmaccent}{\bfseries\textbullet}\hspace{0.48em}#1}
\newtcolorbox{casepanel}{%
  enhanced, colback=casefill, colframe=casestroke, boxrule=0.4pt, arc=6pt,
  boxsep=0pt, left=5pt,right=5pt,top=3pt,bottom=3pt, before skip=0pt,after skip=0pt}
\newtcolorbox{casestatusblock}{%
  enhanced, colback=casebluefill, colframe=casestroke, boxrule=0.3pt, arc=3pt,
  boxsep=0pt, left=6pt,right=6pt,top=3pt,bottom=3pt,
  borderline west={2.2pt}{0pt}{casebluebar}}
\newtcolorbox{caserubricblock}{%
  enhanced, colback=casebluefill, colframe=casestroke, boxrule=0.3pt, arc=3pt,
  boxsep=0pt, left=5pt,right=5pt,top=2.5pt,bottom=2.5pt,
  borderline west={2.2pt}{0pt}{casebluebar}}
\newtcolorbox{caseevidenceblock}{%
  enhanced, colback=casewarmfill, colframe=casewarmstroke, boxrule=0.3pt, arc=3pt,
  boxsep=0pt, left=5pt,right=5pt,top=2.5pt,bottom=2.5pt,
  borderline west={2.2pt}{0pt}{casewarmaccent}}

\newcommand{\casecard}[6]{%
  \par\vspace{0.18em}
  \begin{figure}[H]
    \centering
    \begin{minipage}{0.90\linewidth}
    \centering
    \setlength{\parskip}{0pt}
    \includegraphics[width=\linewidth]{\caseassetdir/#1}\par
    \nointerlineskip
    \begin{casepanel}
        \color{casetext}
        \fontfamily{qhv}\selectfont
        \setlength{\parskip}{0pt}
        \fontsize{7.0}{8.4}\selectfont
        \summarylabel{Verification summary}\par
        \vspace{0.14em}
        \begin{casestatusblock}
          {#3}
        \end{casestatusblock}
        \vspace{0.16em}
        \noindent
        \begin{minipage}[t]{0.40\linewidth}
          \vspace{0pt}
          \begin{caserubricblock}
            \raggedright
            \summarylabel{Rubric}\par
            \vspace{0.08em}
            #4\par
          \end{caserubricblock}
        \end{minipage}
        \hfill
        \begin{minipage}[t]{0.57\linewidth}
          \vspace{0pt}
          \begin{caseevidenceblock}
            \raggedright
            \summarylabel{Evidence}\par
            \vspace{0.04em}
            #5\par
          \end{caseevidenceblock}
        \end{minipage}
    \end{casepanel}
    \end{minipage}
    \captionsetup{font=footnotesize, labelfont=bf, justification=raggedright,
      singlelinecheck=false, skip=0.38em, margin=0pt}
    \caption{#6\label{#2}}
  \end{figure}
  \par\vspace{0.18em}
}

\newfloat{prompt}{htbp}{lop}[section]
\floatname{prompt}{Prompt}

\titlespacing*{\paragraph}{0pt}{1.5ex plus 0.5ex minus 0.2ex}{0.75em}
\title{VeriCUA: Verification as Agentic Exploration for Computer-Using Agents}

\author[1,2*]{Qiushi~Sun}
\author[1]{Da~Yu}
\author[1]{Alexander~Knop}
\author[1]{Amer~Sinha}
\author[2]{Lingpeng~Kong}
\author[2]{Ben~Kao}
\author[1]{Ravi~Kumar}
\author[1]{Pritish~Kamath}
\affil[1]{Google}
\affil[2]{The University of Hong Kong}
\correspondingauthor{qiushisun@connect.hku.hk, pritishk@google.com}
\renewcommand{\copyrightext}{\footerfont * This work was done during Qiushi's internship at Google.}
\renewcommand{\today}{}
\hypersetup{
  pdftitle={VeriCUA: Verification as Agentic Exploration for Computer-Using Agents},
  pdfauthor={Qiushi Sun, Da Yu, Alexander Knop, Amer Sinha, Lingpeng Kong, Ben Kao, Ravi Kumar, Pritish Kamath},
  pdfsubject={Technical Report V0}
}

\begin{abstract}
Building and scaling computer-using agents (CUAs) across desktop, mobile,
and web environments requires reliable task-completion verification for
data curation, evaluation, and reinforcement learning. Typical automated
approaches prompt vision-language models (VLMs) to judge trajectories or
generate programs to inspect application state. However, both rely on
partial views of complex digital environments. VLM judges are restricted
to supplied observations, while generated programs depend on accessible
state. These views can omit decisive evidence or leave completion
judgments unsupported. Tailoring separate verifiers to different platforms and access
settings adds further engineering cost.
We present VeriCUA, a platform-agnostic framework for verification through
agentic exploration.
It treats verification as dynamic evidence gathering and assessment within a
shared procedure for online and offline settings across platforms and
observation modalities.
After inspecting the available context to construct a task rubric, the
verifier selects GUI or code checks for unresolved requirements and combines
their findings under explicit evidence rules.
Experiments show improved agreement with reference labels over the compared
baselines across these settings, with the largest offline gains on
hard cases and stronger rejection of false completion claims online. Further
analyses connect these gains to complementary evidence and test-time
exploration.
\end{abstract}

\begin{document}
\maketitle

\section{Introduction}
\label{sec:intro}

Vision-language models (VLMs) have enabled computer-using agents (CUAs) to
navigate the digital world through software interfaces~\citep{OSWorld,
webarena, androidworld}. Given a natural-language instruction, an agent
observes the interface, plans, and acts, leaving a trajectory of thoughts,
actions, and observations. Digital tasks span applications in which success
can depend on visual presentation, application records, or changes to
persisted files. A trajectory can appear successful even when edits remain
unsaved or an operation fails without a visible warning. The agent may also
report completion after fulfilling only part of the instruction. Accepting
such a run as successful produces a false positive, making the distinction
between attempted actions and their actual effects central to verification.
Developing these agents therefore requires reliable judgments throughout
evaluation, data curation~\citep{sun2025osgenesis}, and reinforcement
learning~\citep{xue2026acurl}, including applications for which no
task-specific checker has been supplied.

An execution can be verified online from the resulting machine state or
offline from its retained trajectory. Evidence can come from rendered
interfaces, structured observations, or artifacts such as files and database
records. VLM judges assess trajectories from selected
observations~\citep{pan2024autonomous}, while automatically generated checkers
inspect application state through model-written
programs~\citep{xue2026evocuaevolvingcomputeruse}. A fixed set of frames may
omit a decisive effect, and a generated query may target state that does not
answer the verification question. Platform-native signals support more
specialized judgments and validation~\citep{chae2026webshepherd,
sun2025sentinel}, but their availability varies with the application and
access setting. Agentic approaches actively inspect trajectories or
environments~\citep{wan2026seekjudge, cui2026vagen, sanayei2026evidencefirst}.
A shared procedure must seek relevant evidence under these access constraints
and ground its conclusions in the observations.

\begin{figure}[t]
\centering
\includegraphics[width=0.95\textwidth]{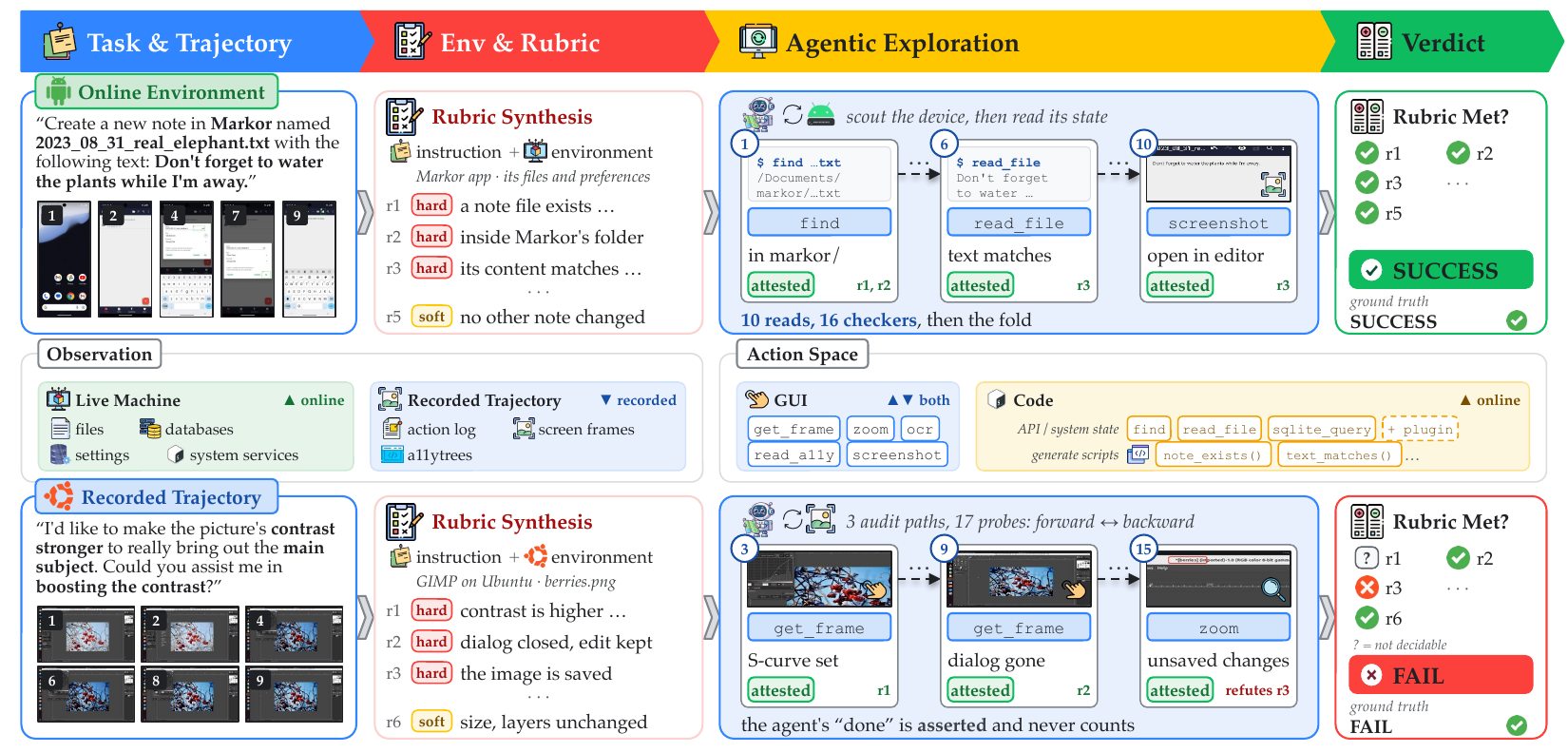}
\caption{VeriCUA inspects the available environment or recording before
constructing a rubric, then gathers evidence to verify its requirements.}
\label{fig:pipeline}
\end{figure}

\Needspace{3\baselineskip}
To address these evidence and access limitations, we present VeriCUA, a
platform-agnostic framework that formulates CUA verification as agentic exploration (Figure~\ref{fig:pipeline}). It provides a shared
procedure for offline and online verification using the sources and inspection
tools available in each setting. Verification begins after the executing
agent's attempt ends. The verifier first explores the available context to
understand the app and its task artifacts, then uses these
observations with the instruction to construct a rubric of fine-grained
requirements. This context helps interpret task concepts and locate potential
evidence, especially when the application is unfamiliar. The verifier chooses
GUI or code tools to check the rubric's requirements. Findings from these
checks reveal what remains unresolved and guide further inspection of relevant
frames, artifacts, or another available source.

VeriCUA's unified design separates the verification workflow from the
interfaces through which evidence is obtained. Offline verification navigates
retained screen frames and action logs, using structured observations such as
accessibility trees when recorded. Online verification can inspect the
accessible machine state left by the executing agent through GUI or code
tools, including files, databases, and system services. These sources answer
different questions about the requested outcome. An action log records an
attempted edit, while a file read can show whether the edited content
persisted. A report that the task is complete remains a claim to verify. The
framework links findings to their supporting observations and records source
properties such as frame timing or state persistence. Explicit evidence rules
combine the findings into a task verdict while retaining which requirements
are supported, refuted, or unresolved.

We evaluate this shared procedure in offline and online scenarios on multiple
platforms. VeriCUA achieves higher agreement with reference labels than the
compared baselines in both settings. It also rejects more false completion
claims than the VLM judge in the online study. The action-space study shows
that adding code-based state reads enables more decisions without increasing
false positives. At a matched number of verification passes, adding visual
evidence improves agreement and decision coverage compared with repeating the
same configuration.

\newpage
Our contributions are summarized below:

\Needspace{7\baselineskip}
\begin{itemize}[leftmargin=1.3em, itemsep=2pt, topsep=2pt, parsep=0pt]
    \item General-Purpose Agentic Verification. VeriCUA formulates CUA
    verification as agentic exploration, providing a shared paradigm for
    offline and online tasks across platforms and available evidence sources.
    \item Adaptive Evidence Gathering. Context inspection grounds task
    rubrics in the available environment or recording. The verifier selects
    GUI or code checks for unresolved requirements and combines findings
    under source-specific evidence rules.
    \item Evaluation Across Platforms. Experiments show improved agreement
    with reference labels across the evaluated platforms. Analyses show how
    evidence access and complementary exploration affect decision coverage
    and reliability.
\end{itemize}
\section{Related Work}
\label{sec:related}

\paragraph{Computer-Using Agents and Their Evaluation.}
Computer-using agents are evaluated on web, mobile, and desktop
tasks~\citep{webarena, androidworld, OSWorld}, with dedicated environments
covering scientific applications~\citep{sun2026scienceboard}. These
benchmarks typically use task-specific programmatic evaluators to check
the resulting application state. Human-labeled trajectory datasets provide
reference judgments across applications~\citep{sun2026osreward}, while
audits of benchmark evaluators identify errors in the checks used to
produce reference labels~\citep{osworldverified}. Benchmarks of
environment-aware judges assess evidence acquisition and verification of
states and execution processes~\citep{shi2026ajbench}. Completion judgments
also support data curation and training beyond benchmark evaluation.
Trajectory synthesis pipelines use verifier scores to select
examples~\citep{sun2025osgenesis}, and continual reinforcement learning
relies on automated feedback during environment
adaptation~\citep{xue2026acurl}. Rubric-based reward modeling further uses
synthetic criteria derived from preferred and rejected
responses~\citep{liu2026openrubrics}. Safety-oriented evaluation additionally
examines whether agent behavior satisfies workflow
constraints~\citep{sun2025sentinel}. The range of applications and
evaluation criteria creates demand for verification that can locate
relevant evidence through available interfaces, including tasks without a
dedicated evaluator.

\paragraph{Verifying Trajectories and Reward Models.}
VLM judges assess task completion from trajectory
observations~\citep{pan2024autonomous}, while structured critics evaluate
task requirements or intermediate milestones~\citep{li2026osthemis}.
Task-adaptive rubrics can be generated from task descriptions to guide
trajectory evaluation~\citep{ding2026adarubric}.
Generated verifiers construct programs that inspect application
state~\citep{xue2026evocuaevolvingcomputeruse}. Reward benchmarks evaluate
the reliability of trajectory judgments against reference labels across tasks and
platforms~\citep{sun2026osreward, lin2025cuarewardbench}. Agent-based
evaluators assess intermediate task requirements in code-generation
workflows~\citep{zhuge2025agentasajudge}. Agentic approaches
gather additional evidence by iteratively inspecting recorded
trajectories~\citep{wan2026seekjudge} or interacting with the execution
environment~\citep{cui2026vagen}. Evidence-first verification separates
investigation from validation, checking proposed claims against their
state or trace sources before accepting rubric
verdicts~\citep{sanayei2026evidencefirst}. These approaches motivate both
seeking task-relevant evidence and assessing how it supports a judgment.
VeriCUA builds its rubric after inspecting the available environment or
recording, then selects GUI or code checks for unresolved requirements.
Source-aware rules combine the resulting findings within a shared offline
and online verification workflow.

\section{Method}
\label{sec:method}

\subsection{Verification as Active Evidence Gathering}
\label{sec:formulation}

Given an instruction $g$, an executing agent produces a trajectory
$\tau = [o_0, a_0, \ldots, o_T]$ of observations and actions, together with
any recorded reasoning, and leaves the environment in state $s_T$.
VeriCUA begins after this attempt ends and assesses whether the execution
fulfilled $g$ using the sources available to the verifier. Although task
completion is binary, the verifier outputs
$y \in \{\textsc{success}, \textsc{fail}, \textsc{abstain}\}$, allowing
abstention when the gathered evidence is insufficient to support a decision.

\Needspace{3\baselineskip}
The available sources $\mathcal{S} = \{S_1, \ldots, S_m\}$ can include a
retained trajectory, recorded accessibility trees (\texttt{a11ytrees}), a frozen
machine snapshot, or accessible post-execution machine state. Offline
verification uses retained observations, while online verification can
inspect exposed machine state alongside any retained execution records.
When a requirement concerns a change made during execution, an available
pre-execution reference can also serve as a source.

The shared workflow in Figure~\ref{fig:pipeline} begins with \emph{context
inspection}, which gathers information about the application and available
observations into context $c$. \emph{Rubric synthesis} uses $g$
and $c$ to produce requirements $R$ that make the instruction checkable.
During \emph{active exploration}, the verifier selects checks for these
requirements and uses their results to decide what to inspect next.
Finally, \emph{rule-based arbitration} combines the findings under the
setting's evidence rules and handles any unresolved requirements.

At step $k$, an inspection action
$\alpha_k = (v_k, x_k) \in \mathcal{A}(\mathcal{S})$ pairs a tool $v_k$
with arguments $x_k$. Let $\mathcal{E}_{k-1}$ denote the inspection record
before this step, initialized with any retained context-inspection outputs
$\mathcal{E}_0$. The verifier selects and executes the next check according to
\begin{equation}
\alpha_k \sim \pi_V(\,\cdot \mid g, c, R, \mathcal{E}_{k-1}),
\qquad
\operatorname{Execute}(\alpha_k,\mathcal{S})
    \longrightarrow (\tilde{o}_k,t_k),
\label{eq:verification-policy}
\end{equation}
where $\tilde{o}_k$ is the returned tool observation and $t_k$ records its
evidence type and source attributes, assigned by the tool-execution layer
(the \emph{harness}). The executed check
$(\alpha_k,\tilde{o}_k,t_k)$ and its associated findings are added to form
$\mathcal{E}_k$. A \emph{finding} assesses how an observation supports or
refutes a particular rubric requirement. The verifier continues until the
hard requirements are settled or its inspection budget is exhausted. After
$L$ inspection steps, the arbiter derives a verdict from $\mathcal{E}_L$.

\subsection{Environment-Aware Rubric Synthesis}
\label{sec:decompose}

An instruction specifies the requested result but may leave its
application-specific interpretation implicit. A request to save a note,
for example, does not necessarily identify the application's storage
directory or explain which view reveals whether the note persisted.
Generating requirements without this context can introduce assumptions
about files or interface elements that the application does not expose.

Before constructing the rubric, VeriCUA uses the available inspection
tools to gather context $c$. Online inspection can identify application
packages, relevant directories, and exposed database schemas; offline
inspection reviews the recorded steps, frames, and available structured
observations. From the instruction $g$ and observed context $c$, the verifier
constructs a rubric $R = \{(r_i,h_i)\}_{i=1}^{n}$, where each $r_i$ is a
checkable task requirement and
$h_i \in \{\text{hard},\text{soft}\}$ indicates whether it is mandatory
for success or an additional check. Requirements can concern a resulting
state or a change relative to an available pre-execution reference.

Each hard requirement must be justified by the outcomes requested in the
instruction, while context supplies its application-specific meaning and
helps locate evidence. In the note-saving
example, the requested text and persistence are task requirements; a
discovered directory is a place to inspect for evidence. This distinction
keeps evidence locations from becoming additional conditions for success.
The rubric guides subsequent checks, which assign each requirement a
status of \textsc{met}, \textsc{unmet}, or \textsc{unknown}. Soft checks provide additional findings
without relaxing any hard requirement.

\subsection{Adaptive Hybrid Action Space}
\label{sec:action-space}

\begin{figure}[t]
\centering
\includegraphics[width=\textwidth]{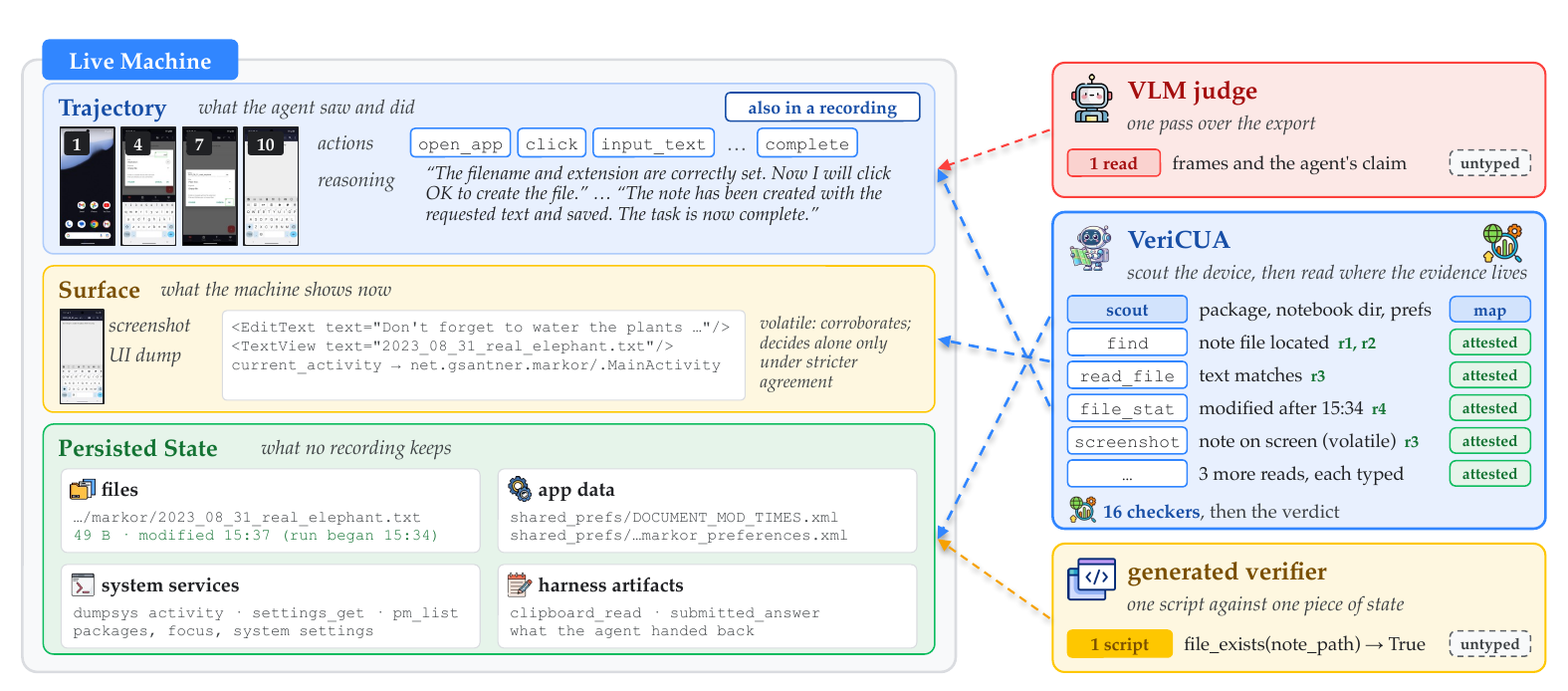}
\caption{VeriCUA gathers evidence from retained trajectories and, when
accessible, the current interface and underlying machine state. Each read
retains its source attributes.}
\label{fig:machine}
\end{figure}

The action space connects the rubric to the sources that can answer its
requirements. Figure~\ref{fig:machine} distinguishes three views of task
state:
\begin{itemize}[leftmargin=1.3em, itemsep=2pt, topsep=3pt, parsep=0pt]
    \item \emph{Trajectory.} Retained frames, actions, and reasoning record
    what the executing agent observed and attempted. The verifier can
    search the action log, inspect selected frames, and read recorded
    \texttt{a11ytrees} when available.
    \item \emph{Surface.} The current machine interface can be inspected
    through a new screenshot or a UI hierarchy read. A recorded screenshot
    belongs to the historical trajectory; this view provides a fresh
    observation of the accessible machine.
    \item \emph{Persisted State.} Files, application databases, and stored
    settings can retain effects that the current interface does not show.
    Code tools can inspect them through file reads, database queries, or
    queries to exposed services.
\end{itemize}
The harness records persistence for each read. Alongside stored artifacts,
live machine access provides volatile observations such as current UI
dumps and harness artifacts.

GUI tools support visual inspection through frame selection, region
enlargement, and optical character recognition (OCR), while code tools
query exposed machine state. The online verifier can also
synthesize a \emph{checker}: a sequence of permitted reads paired with a
predicate that evaluates a requirement. The harness executes the reads;
structured predicates use a restricted evaluator, while visual or semantic
comparisons use a model judgment. A single online inspection action
$\alpha_k$ can execute a checker comprising several reads and its predicate
evaluation.

The verifier chooses its next inspection from the findings already
collected, focusing on requirements that remain unresolved within the
available access. A text match may leave persistence unresolved and prompt
a file read; an unclear frame may prompt a crop or inspection of a later
frame. The harness records the executed
tool, arguments, and source metadata, including frame timing or the
persistence of queried state. The executed read determines these
attributes, which are attached to the resulting findings.
The complete tool inventory is given in
Appendix~\ref{app:impl-actions}.

\subsection{Evidence Discipline}
\label{sec:evidence}

VeriCUA links each finding to an executed read, then applies source
constraints and corroboration rules before assigning requirement statuses.

\paragraph{Grounded Findings.}
Each finding identifies a rubric requirement, states the support or
refutation inferred from an observation, and cites the executed check that
returned the observation. The harness rejects citations that match no executed call and
attaches each valid citation's observation and source attributes to the
finding. Source attributes and corroborating probes then determine how
each linked assessment contributes to a requirement's status.

\paragraph{Source-Aware Evidence.}
The \emph{attested}/\emph{asserted} source labels distinguish observations
of task state from records of the executing agent's intent or claims. A
post-action frame, a file read, or a
database query can provide attested evidence. The executing agent's
completion message and a pre-action frame showing an intended operation
are asserted evidence for that operation's outcome. Asserted evidence can
direct further inspection but cannot alone mark a requirement \textsc{met}.

Source attributes further constrain what an observation can support,
including whether a recorded image precedes or follows the action whose
effect is being checked. Online reads distinguish persistent artifacts
from volatile interface state, which can answer different questions about
the same requirement. For example, a screen displaying
edited text can support its visible contents, while a file read can check
whether that text was saved. A finding must address the task requirement,
and its observation must satisfy the source conditions for the inference
being made.

\paragraph{Rule-Based Arbitration.}
The arbiter applies the setting's confirmation and refutation rules to the
stored findings $\mathcal{E}_L$, assigning each requirement a status of
\textsc{met}, \textsc{unmet}, or \textsc{unknown}. A \textsc{met} status
requires attested support, while \textsc{unmet} requires corroborating
refutations from separate probes. Evidence that does not settle the
requirement leaves it \textsc{unknown}. The task verdict is
\textsc{success} when all hard requirements are \textsc{met}, and
\textsc{fail} when any is \textsc{unmet}. If some remain \textsc{unknown}
and none is \textsc{unmet}, the task is unresolved and follows the
setting's abstention or fallback policy. Aggregation is deterministic for
fixed findings and rules; the stored records allow confirmation thresholds
to be re-evaluated without repeating inspection actions.

\subsection{Online and Offline Instantiations}
\label{sec:instantiations}

Online and offline verification share the same workflow over the sources
illustrated in Figure~\ref{fig:machine}. Access conditions determine the
available checks and the treatment of unresolved tasks.

\paragraph{Online.}
Online machine access allows the verifier to combine GUI inspection with
code-based checks of stored artifacts. Task-specific checkers can be
reused across repeated executions of the same task with compatible
application schemas. Persistent attested reads can confirm requirements,
while volatile observations provide corroboration. The verifier abstains on unresolved
tasks to avoid assigning training or filtering labels unsupported by the
collected evidence.

\paragraph{Offline.}
VeriCUA audits fixed, time-stamped recordings through multiple paths,
seeking support through forward inspection and refutations through
backward inspection. A requirement is \textsc{met} when every completed
path provides attested support. For tasks left unresolved, the reported
configuration uses an \emph{initial judgment}: a single-pass VLM assessment
of the retained trajectory computed before exploration and withheld from
the audit paths. This fallback supplies the final verdict while unresolved requirements remain
\textsc{unknown}. Detailed setting rules and inspection budgets are given
in Appendices~\ref{app:impl-settings} and~\ref{app:exp-config}.

\section{Experiments}
\label{sec:exp}

\subsection{Experimental Setup}
\label{sec:exp_setup}

\paragraph{Benchmarks.}
In online environments, we run AndroidWorld \citep{androidworld} on an Android
emulator, OSWorld \citep{OSWorld} in a virtual machine, and ScienceBoard
\citep{sun2026scienceboard} in its scientific desktop environment. Each
episode is executed by the benchmark's own agent scaffold, and the verifier
probes the resulting machine state after the episode ends. For offline
verification, we evaluate recorded WebArena trajectories from three
executing agents~\citep{webarena} and the human-labeled OSReward
trajectories~\citep{sun2026osreward}, which span four platforms and several
agents.

\paragraph{Protocol.}
All verifiers and baselines run on \texttt{gemini-3-flash-preview}, and the
backbone study re-runs them on \texttt{gemini-3.1-pro-preview}. In online
environments, gold is each benchmark's official evaluator and the executing
agents also run on \texttt{gemini-3-flash-preview}. On WebArena we provide
\texttt{a11ytree} variants of the input beyond the GUI, to the verifier and the judges
alike. The remaining protocol details are provided in Appendix~\ref{app:exp}.

\paragraph{Baselines.}
We compare against a VLM judge, an automatically generated verifier, and two
structured judges. Implementation details are provided
in Appendix~\ref{app:exp-baselines}.

\begin{itemize}[leftmargin=1.3em, itemsep=1pt, topsep=2pt, parsep=0pt]
    \item \textbf{VLM judge.} A single-pass judge such as
    TRM~\citep{sun2025osgenesis}, which assesses the supplied observations
    in one model call and returns a verdict or a score.
    \item \textbf{Auto-Generated Verifier}
    \citep{xue2026evocuaevolvingcomputeruse}. A model-generated evaluator
    script per task, run once against the machine for a binary verdict.
    \item \textbf{CUAJudge} \citep{xue2026acurl}. A checklist judge that derives
    key points from the task and checks them against sampled frames in one
    call.
    \item \textbf{OS-Themis} \citep{li2026osthemis}. A milestone judge that
    checks ordered milestones and reviews them.
\end{itemize}
The VLM judge runs in both settings, the auto-generated verifier only in
online verification, and CUAJudge and OS-Themis only in offline
verification.

\paragraph{Metrics.}
We report agreement with gold, precision, recall, and F1 for the \textsc{success}
class, and the false positive rate, with abstentions counted as \textsc{fail}. For
verifiers that abstain we also report the decision rate and the accuracy on
decided episodes. Verdict conventions are given in Appendix~\ref{app:exp-metrics}.

\subsection{Online Verification}
\label{sec:results-live}

\begin{table}[t]
\centering
\small
\caption{Performance in online environments.}
\label{tab:live}
\setlength{\tabcolsep}{2.2pt} %
\renewcommand{\arraystretch}{1.15}
\begin{tabular}{@{}llccccccc@{}}
\toprule
\textbf{Environment} & \textbf{Judge} & \textbf{Prec} & \textbf{Rec} & \textbf{F1} & \textbf{Acc} &
\textbf{False Positive Rate} $\downarrow$ & \textbf{Decided} & \textbf{Decided Acc} \\
\midrule
\multirow{3}{*}{AndroidWorld} & VLM judge & 79.3 & 69.7 & 74.2 & 78.4 & 15.4 & 100\% & 78.4\% \\
 & Auto-Generated verifier & 77.8 & 84.8 & 81.2 & 82.4 & 16.9 & 100\% & 82.4\% \\
 & \textbf{VeriCUA} & \textbf{90.6} & \textbf{87.9} & \textbf{89.2} & \textbf{90.5} & \textbf{7.7} & 82\% & \textbf{95.1\%} \\
\midrule
\multirow{3}{*}{OSWorld} & VLM judge & 68.2 & 75.0 & 71.4 & 73.4 & 27.9 & 100\% & 73.4\% \\
 & Auto-Generated verifier & 70.6 & \textbf{78.1} & 74.2 & 75.9 & 25.9 & 100\% & 75.9\% \\
 & \textbf{VeriCUA} & \textbf{80.3} & 71.3 & \textbf{75.5} & \textbf{79.5} & \textbf{13.9} & 78\% & \textbf{88.7\%} \\
\midrule
\multirow{3}{*}{ScienceBoard} & VLM judge & 42.9 & 60.0 & 50.0 & 71.6 & 24.8 & 100\% & 71.6\% \\
 & Auto-Generated verifier & 50.9 & \textbf{72.5} & 59.8 & 76.9 & 21.7 & 100\% & 76.9\% \\
 & \textbf{VeriCUA} & \textbf{77.4} & 60.0 & \textbf{67.6} & \textbf{86.4} & \textbf{5.4} & 70\% & \textbf{91.6\%} \\
\bottomrule
\end{tabular}
\end{table}

\paragraph{Accuracy and False Positives.}
Across the three online environments, VeriCUA has the highest accuracy and
lowest false-positive rate (Table~\ref{tab:live}). The auto-generated
verifier has higher accuracy than the VLM judge in each environment.
The largest gap is on ScienceBoard, where generated checks are not
reliably aligned with the target scientific software. Scripts constructed
without context inspection can query incorrect state locations. VeriCUA
inspects the application context before generating checks, and its
false-positive rate is roughly a quarter of the generated verifier's.
The generated verifier has higher recall on OSWorld and ScienceBoard,
while VeriCUA has higher precision and accuracy.

\paragraph{Deciding and Abstaining.}
Both baselines return a verdict on every episode, so their decided
accuracy equals their overall accuracy. VeriCUA abstains when its findings
cannot settle the task or when decided verification passes disagree. Its accuracy
on decided episodes is higher across all three environments. Counting
abstentions as \textsc{fail} can lower recall without adding false
positives. This policy favors conservative selection of successful
trajectories for data curation or reward assignment.

\subsection{Offline Verification}
\label{sec:results-recorded}

\begin{table}[!ht]
\centering
\small
\caption{Performance in offline verification.}
\label{tab:recorded}
\setlength{\tabcolsep}{3.4pt}
\renewcommand{\arraystretch}{1.15}
\begin{tabular}{@{}lcccccccccc@{}}
\toprule
\multirow{2}{*}{\textbf{Judge}} & \multicolumn{2}{c}{\textbf{OSReward-Hard}} & \multicolumn{2}{c}{\textbf{OSReward}} & \multicolumn{2}{c}{\textbf{WebArena}} & \multicolumn{2}{c}{\textbf{AndroidWorld}} & \multicolumn{2}{c}{\textbf{OSWorld}} \\
\cmidrule(lr){2-3} \cmidrule(lr){4-5} \cmidrule(lr){6-7} \cmidrule(lr){8-9} \cmidrule(lr){10-11}
 & Acc & BalAcc & Acc & BalAcc & Acc & BalAcc & Acc & BalAcc & Acc & BalAcc \\
\midrule
TRM & 57.4 & 64.8 & 82.2 & 83.1 & 79.6 & 79.8 & 88.3 & 89.6 & 80.3 & 73.7 \\
CUAJudge & 63.4 & 62.5 & 83.7 & 82.6 & 81.3 & 81.8 & 91.2 & 90.7 & 83.4 & 78.2 \\
OS-Themis & 65.7 & 63.2 & 80.2 & 79.8 & 75.4 & 74.3 & 89.4 & 89.0 & 82.1 & 79.0 \\
\textbf{VeriCUA} & \textbf{70.4} & \textbf{68.3} & \textbf{87.2} & \textbf{86.3} & \textbf{82.7} & \textbf{83.1} & \textbf{94.3} & \textbf{94.1} & \textbf{86.3} & \textbf{81.2} \\
\bottomrule
\end{tabular}
\end{table}

\paragraph{Across Benchmarks and Difficulty.}
In offline verification, VeriCUA has the highest accuracy and balanced
accuracy across the benchmarks in Table~\ref{tab:recorded}. On OSReward,
its accuracy advantage over the strongest compared judge is larger on the
hard subset than on the full set. The gains on hard trajectories support
active evidence gathering when task completion is difficult to assess
from fixed observations.

\paragraph{Robustness and Evidence Limits.}
On WebArena, VeriCUA remains ahead across the three rollout agents and
when baseline judges receive accessibility trees alone. The per-agent and
observation comparisons are reported in
Appendix~\ref{app:ana-observation}. Offline verification is limited by the
observations retained in the recording. When these do not resolve the
task, VeriCUA uses its initial-judgment fallback. This occurs on 23\% of
WebArena trajectories. Appendix~\ref{app:cases} presents online and offline
examples with the evidence underlying their verdicts.

\section{Analysis}
\label{sec:analysis}

\subsection{GUI Versus Hybrid Action Space}
\label{sec:ana-action}

\begin{figure}[b]
\centering
\begin{minipage}[c]{0.44\textwidth}
\centering
\includegraphics[width=\textwidth]{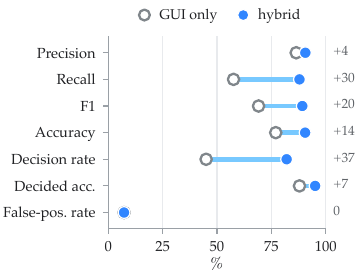}
\end{minipage}\hfill
\begin{minipage}[c]{0.52\textwidth}
\caption{Each metric of VeriCUA with GUI actions alone and with the hybrid
action space on the online AndroidWorld suite. Hollow markers are GUI only,
filled markers the hybrid space, and the column at the right gives the
change.}
\label{fig:action}
\end{minipage}
\end{figure}
\paragraph{GUI Only versus Hybrid.}
On online AndroidWorld, adding code-based state checks to GUI inspection
increases the decision rate and recall while leaving false positives
unchanged (Figure~\ref{fig:action}). In the paired analysis, every
additional correct decision from the code channel is supported by a file
or application-store read. Access to stored state therefore resolves
requirements that GUI observations leave unsettled.

\subsection{The Effect Of Observation}
\label{sec:ana-obs}

\begin{figure}[t]
\centering
\includegraphics[width=0.9\textwidth]{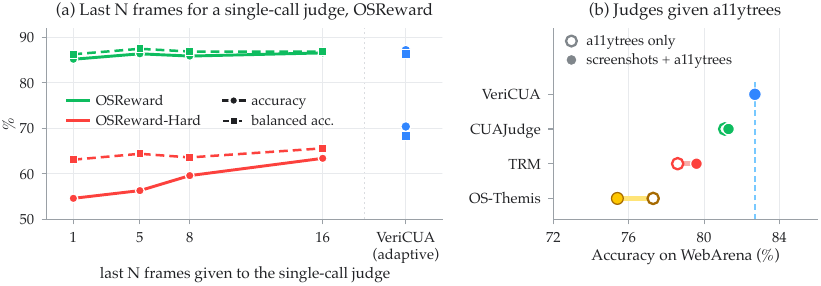}
\vspace{-6pt} %
\caption{More observation for a single-call judge. (a) The OSReward judge
reading the last $N$ frames of each trajectory, with VeriCUA at the right;
OSReward's default is five. (b) The WebArena judges reading \texttt{a11ytrees} alone
or with screenshots, against VeriCUA (dashed).}
\label{fig:obs}
\end{figure}

\paragraph{Different Observations.}
On WebArena, adding \texttt{a11ytrees} to screenshots raises VeriCUA's
decision rate while leaving accuracy unchanged. Trees alone yield the
highest decision rate and lowest cost, at lower accuracy.
Table~\ref{tab:obs_input} in Appendix~\ref{app:ana-observation} reports the
observation-input ablation. The comparison with baseline judges uses
trajectories from three rollout agents. Their accuracy changes little
between tree-only and combined inputs, and VeriCUA remains ahead under
both settings (Figure~\ref{fig:obs}b). Per-agent results are given in
Table~\ref{tab:webarena_tree}.

\paragraph{More Observations.}
We also vary the number of final frames given to the single-call
OSReward judge (Figure~\ref{fig:obs}a). Additional frames improve its
accuracy on the hard subset, but the judge remains below VeriCUA.
Accuracy on the full set changes only modestly.

\subsection{Test-Time Exploration}
\label{sec:ana-tts}

\begin{figure}[b]
\centering
\includegraphics[width=\textwidth]{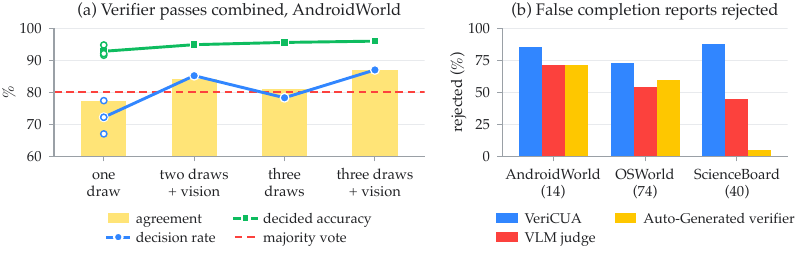}
\vspace{-6pt} %
\caption{(a) Combining verifier passes improves agreement and decision rate.
(b) Rejection of the agent's false completion reports.}
\label{fig:multipass}
\end{figure}

\paragraph{Combining Passes.}
On the multi-pass AndroidWorld suite, three passes use the default
configuration of code and structured-state checks. An additional pass
enables visual inspection through screenshots and file-image reads,
referred to as the vision fallback. All passes and the auto-generated
baseline share one restored snapshot per episode. The combination rule
retains a decided verdict and abstains when decided passes disagree.
Combining the default passes gives modest gains over the strongest single
pass. Adding the visual pass improves agreement and decision rate further,
both reaching 87\% (Figure~\ref{fig:multipass}a). The visual pass also
improves these metrics when the number of passes is matched. Combined
verdicts produce fewer incorrect \textsc{fail} decisions than majority
voting over the same passes
(Table~\ref{tab:multipass}, Appendix~\ref{app:ana-tts}).

\begin{wraptable}{r}{0.46\textwidth}
\vspace{-6pt}
\centering
\small
\caption{Accuracy under matched budgets.}
\label{tab:budget}
\setlength{\tabcolsep}{4pt}
\renewcommand{\arraystretch}{1.1}
\begin{tabular}{@{}lcc@{}}
\toprule
\textbf{Judge} & \textbf{OSReward} & \textbf{OSReward-Hard} \\
\midrule
TRM (N=10) & 83.5 & 58.5 \\
CUAJudge & 84.8 & 64.5 \\
\textbf{VeriCUA} & \textbf{86.4} & \textbf{68.7} \\
\bottomrule
\end{tabular}
\vspace{-12pt}
\end{wraptable}
\paragraph{Matched Budget.}
We also compare accuracy under a matched verification budget on the flash
backbone (Table~\ref{tab:budget}). The protocol allocates repeated runs to
TRM, a fixed step limit to CUAJudge, and an exploration limit to VeriCUA.
VeriCUA retains higher accuracy on OSReward and OSReward-Hard. Restricting
its budget produces a small accuracy decrease from the default
configuration while preserving the lead over the compared judges.

\paragraph{Detecting False Completion Reports.}
We analyze false completion reports across the three online environments,
using an AndroidWorld subset (Figure~\ref{fig:multipass}b).
VeriCUA rejects about 80\% of these reports and
outperforms both baselines in each environment. The difference is largest
on ScienceBoard, where the generated verifier rarely rejects an incorrect
completion claim. Figure~\ref{fig:case-camera} in
Appendix~\ref{app:cases} illustrates an AndroidWorld case in which reads of
the media store refute the agent's report.

\subsection{Gains From A Stronger Backbone}
\label{sec:ana-backbone}

\begin{wraptable}{r}{0.55\textwidth}
\vspace{-12pt}
\centering
\small
\caption{Backbone comparison on OSReward-Hard.}
\label{tab:backbone}
\setlength{\tabcolsep}{3pt}
\renewcommand{\arraystretch}{1.1}
\begin{tabular}{@{}lcccccc@{}}
\toprule
\multirow{2}{*}{\textbf{Judge}} & \multicolumn{3}{c}{\textbf{Acc}} & \multicolumn{3}{c}{\textbf{BalAcc}} \\
\cmidrule(lr){2-4} \cmidrule(lr){5-7}
 & \textbf{Flash} & \textbf{Pro} & $\Delta$ & \textbf{Flash} & \textbf{Pro} & $\Delta$ \\
\midrule
CUAJudge & 63.4 & 62.7 & $-$0.7 & 62.5 & 58.4 & $-$4.1 \\
OS-Themis & 65.7 & 66.1 & +0.4 & 63.2 & 61.1 & $-$2.1 \\
\textbf{VeriCUA} & \textbf{70.4} & \textbf{73.1} & \textbf{+2.7} & \textbf{68.3} & \textbf{69.2} & \textbf{+0.9} \\
\bottomrule
\end{tabular}
\vspace{-8pt}
\end{wraptable}

Replacing the flash backbone with the pro backbone improves VeriCUA's
accuracy and balanced accuracy on OSReward-Hard
(Table~\ref{tab:backbone}). The compared judges show smaller or negative
changes, indicating that stronger models benefit the verification methods
differently. All frameworks improve in the online AndroidWorld comparison,
with VeriCUA retaining the highest accuracy on both backbones
(Table~\ref{tab:backbone_online}, Appendix~\ref{app:ana-backbone}).

\FloatBarrier
\subsection{Confident Failures and Gold-Label Audit}
\label{sec:ana-audit}

\begin{wrapfigure}{r}{0.52\textwidth}
\centering
\includegraphics[width=\linewidth]{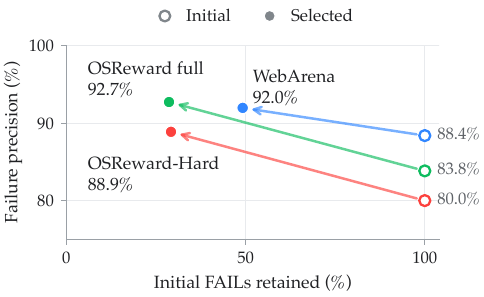}
\caption{Failure precision and retention.}
\label{fig:failure-precision}
\end{wrapfigure}

For offline verification, we compare initial \textsc{fail} judgments with the
subset that exploration also resolves as \textsc{fail}. Selection raises failure
precision on each benchmark while retaining fewer of the initial failure
judgments (Figure~\ref{fig:failure-precision}).

Disagreements can also identify suspected errors in the reference labels.
A blinded model adjudicator supports VeriCUA on 63\% of the flagged
trajectories, with the clearest support on OSWorld. These results suggest
that verification can help identify failed trajectories and prioritize
suspected label errors for inspection. Appendix~\ref{app:ana-audit} details
the supporting analyses.
\FloatBarrier

\section{Conclusion}
\label{sec:conclusion}

We presented \ours, a platform-agnostic framework for verifying
computer-using agents through agentic exploration. The verifier inspects
available context before constructing a task rubric, then selects GUI or
code checks to gather evidence for unresolved requirements. Source-aware
arbitration combines the findings within a shared workflow for offline
recordings and online environments. Across the evaluated platforms, \ours
improves agreement with reference labels over the compared baselines, with
fewer false positives online and the largest offline gains on hard
trajectories. Analyses show that complementary evidence sources and
verification passes improve decision coverage, while stronger backbones
further improve accuracy. These results support agentic verification for
CUA evaluation and motivate its use in data curation and reinforcement
learning.

\bibliographystyle{abbrvnat}
\bibliography{references}

\begin{thebibliography}{21}
\providecommand{\natexlab}[1]{#1}
\providecommand{\url}[1]{\texttt{#1}}
\expandafter\ifx\csname urlstyle\endcsname\relax
  \providecommand{\doi}[1]{doi: #1}\else
  \providecommand{\doi}{doi: \begingroup \urlstyle{rm}\Url}\fi

\bibitem[Chae et~al.(2025)Chae, Kim, Cho, Kim, Moon, Hwangbo, Lim, Kim, Hwang,
  Gwak, Choi, Kang, Im, Cho, Kim, Han, Kwon, Kim, woo Kwak, Kang, and
  Yeo]{chae2026webshepherd}
H.~Chae, S.~Kim, J.~Cho, S.~Kim, S.~Moon, G.~Hwangbo, D.~Lim, M.~Kim, Y.~Hwang,
  M.~Gwak, D.~Choi, M.~Kang, G.~Im, B.~Cho, H.~Kim, J.~H. Han, T.~Kwon, M.~Kim,
  B.~woo Kwak, D.~Kang, and J.~Yeo.
\newblock Web-shepherd: Advancing {PRM}s for reinforcing web agents.
\newblock In \emph{The Thirty-ninth Annual Conference on Neural Information
  Processing Systems}, 2025.
\newblock URL \url{https://openreview.net/forum?id=G2kMroO9UV}.

\bibitem[Cui et~al.(2026)Cui, Huang, Wang, Zheng, Kong, and Zeng]{cui2026vagen}
C.~Cui, J.~Huang, S.~Wang, L.~Zheng, Q.~Kong, and Z.~Zeng.
\newblock Agentic reward modeling: Verifying gui agent via progressive
  trajectory-grounded interaction, 2026.
\newblock URL \url{https://arxiv.org/abs/2602.00575}.

\bibitem[Ding(2026)]{ding2026adarubric}
L.~Ding.
\newblock Adarubric: Task-adaptive rubrics for reliable llm agent evaluation
  and reward learning.
\newblock 2026.
\newblock URL \url{https://github.com/alphadl/AdaRubrics}.

\bibitem[Li et~al.(2026)Li, Wu, Zhao, Yang, Xie, Liu, Liu, Jin, Liang, Li, Wu,
  Zhou, Wang, and Ding]{li2026osthemis}
Z.~Li, Z.~Wu, Y.~Zhao, B.~Yang, J.~Xie, Z.~Liu, Z.~Liu, K.~Jin, J.~Liang,
  Z.~Li, F.~Wu, B.~Zhou, Z.~Wang, and Z.~Ding.
\newblock Os-themis: A scalable critic framework for generalist gui rewards,
  2026.
\newblock URL \url{https://arxiv.org/abs/2603.19191}.

\bibitem[Lin et~al.(2025)Lin, Tan, Qin, Xu, Shi, Li, Li, Cai, Cai, Fu, Li, and
  Sun]{lin2025cuarewardbench}
H.~Lin, X.~Tan, Y.~Qin, Z.~Xu, Y.~Shi, Z.~Li, G.~Li, S.~Cai, S.~Cai, C.~Fu,
  K.~Li, and X.~Sun.
\newblock Cuarewardbench: A benchmark for evaluating reward models on
  computer-using agent, 2025.
\newblock URL \url{https://arxiv.org/abs/2510.18596}.

\bibitem[Liu et~al.(2026)Liu, Xu, Yu, Hong, Yang, Zhao, and
  Wang]{liu2026openrubrics}
T.~Liu, R.~Xu, T.~Yu, I.~Hong, C.~Yang, T.~Zhao, and H.~Wang.
\newblock {O}pen{R}ubrics: Towards scalable synthetic rubric generation for
  reward modeling and {LLM} alignment.
\newblock In \emph{Proceedings of the 64th Annual Meeting of the {A}ssociation
  for {C}omputational {L}inguistics (Volume 1: Long Papers)}, pages
  17417--17437, San Diego, California, United States, July 2026. Association
  for Computational Linguistics.
\newblock ISBN 979-8-89176-390-6.
\newblock \doi{10.18653/v1/2026.acl-long.791}.
\newblock URL \url{https://aclanthology.org/2026.acl-long.791/}.

\bibitem[Pan et~al.(2024)Pan, Zhang, Tomlin, Zhou, Levine, and
  Suhr]{pan2024autonomous}
J.~Pan, Y.~Zhang, N.~Tomlin, Y.~Zhou, S.~Levine, and A.~Suhr.
\newblock Autonomous evaluation and refinement of digital agents.
\newblock In \emph{First Conference on Language Modeling}, 2024.
\newblock URL \url{https://openreview.net/forum?id=NPAQ6FKSmK}.

\bibitem[Rawles et~al.(2025)Rawles, Clinckemaillie, Chang, Waltz, Lau, Fair,
  Li, Bishop, Li, Campbell-Ajala, Toyama, Berry, Tyamagundlu, Lillicrap, and
  Riva]{androidworld}
C.~Rawles, S.~Clinckemaillie, Y.~Chang, J.~Waltz, G.~Lau, M.~Fair, A.~Li, W.~E.
  Bishop, W.~Li, F.~Campbell-Ajala, D.~K. Toyama, R.~J. Berry, D.~Tyamagundlu,
  T.~P. Lillicrap, and O.~Riva.
\newblock Androidworld: A dynamic benchmarking environment for autonomous
  agents.
\newblock In \emph{The Thirteenth International Conference on Learning
  Representations}, 2025.
\newblock URL \url{https://openreview.net/forum?id=il5yUQsrjC}.

\bibitem[Sanayei et~al.(2026)Sanayei, Dumitru, Hou, Rezaei, Riaz, Vacareanu,
  Yang, Liu, Surdeanu, and Zhang]{sanayei2026evidencefirst}
R.~Sanayei, R.-G. Dumitru, S.~Hou, M.~Rezaei, H.~Riaz, R.~Vacareanu, M.~Yang,
  Y.~Liu, M.~Surdeanu, and D.~Y. Zhang.
\newblock Evidence-first agentic verification, 2026.
\newblock URL \url{https://arxiv.org/abs/2610.17724v1}.

\bibitem[Shi et~al.(2026)Shi, Wang, Zhao, Chen, Feng, Hao, Su, GU, Su, Cai, and
  He]{shi2026ajbench}
W.~Shi, Y.~Wang, Y.~Zhao, Y.~Chen, F.~Feng, X.~Hao, X.~Su, Q.~GU, H.~Su,
  X.~Cai, and X.~He.
\newblock {AJ}-bench: Benchmarking agent-as-a-judge for environment-aware
  evaluation.
\newblock In \emph{Findings of the {A}ssociation for {C}omputational
  {L}inguistics: {ACL} 2026}, pages 25371--25413, San Diego, California, United
  States, July 2026. Association for Computational Linguistics.
\newblock ISBN 979-8-89176-395-1.
\newblock \doi{10.18653/v1/2026.findings-acl.1269}.
\newblock URL \url{https://aclanthology.org/2026.findings-acl.1269/}.

\bibitem[Sun et~al.(2025)Sun, Cheng, Ding, Jin, Wang, Xu, Wu, Jia, Chen, Liu,
  et~al.]{sun2025osgenesis}
Q.~Sun, K.~Cheng, Z.~Ding, C.~Jin, Y.~Wang, F.~Xu, Z.~Wu, C.~Jia, L.~Chen,
  Z.~Liu, et~al.
\newblock Os-genesis: Automating gui agent trajectory construction via reverse
  task synthesis.
\newblock In \emph{Proceedings of the 63rd Annual Meeting of the Association
  for Computational Linguistics (Volume 1: Long Papers)}, pages 5555--5579,
  2025.

\bibitem[Sun et~al.(2026{\natexlab{a}})Sun, Cheng, Wang, Yang, Yan, Chen, Xu,
  Ding, Chen, Cao, et~al.]{sun2026osreward}
Q.~Sun, K.~Cheng, Y.~Wang, B.~Yang, H.~Yan, L.~Chen, F.~Xu, Z.~Ding, N.~Chen,
  J.~Cao, et~al.
\newblock Osreward: Instituting standardized evaluation for cross-platform
  computer-use reward models.
\newblock \emph{arXiv preprint arXiv:2607.28609}, 2026{\natexlab{a}}.

\bibitem[Sun et~al.(2026{\natexlab{b}})Sun, Li, Liu, Xie, Xu, Yin, Cheng, Li,
  Ding, Liu, Wu, Zhang, Kao, and Kong]{sun2025sentinel}
Q.~Sun, M.~Li, Z.~Liu, Z.~Xie, F.~Xu, Z.~Yin, K.~Cheng, Z.~Li, Z.~Ding, Q.~Liu,
  Z.~Wu, Z.~Zhang, B.~Kao, and L.~Kong.
\newblock {OS}-sentinel: Towards safety-enhanced mobile {GUI} agents via hybrid
  validation in realistic workflows.
\newblock In \emph{Proceedings of the 64th Annual Meeting of the {A}ssociation
  for {C}omputational {L}inguistics (Volume 1: Long Papers)}, pages 9529--9553,
  San Diego, California, United States, July 2026{\natexlab{b}}. Association
  for Computational Linguistics.
\newblock ISBN 979-8-89176-390-6.
\newblock URL \url{https://aclanthology.org/2026.acl-long.431/}.

\bibitem[Sun et~al.(2026{\natexlab{c}})Sun, Liu, Ma, Ding, Xu, Yin, Zhao, Wu,
  Cheng, Liu, Wang, Li, Tang, Xie, Feng, Li, Kao, Wang, Qi, Kong, and
  Wu]{sun2026scienceboard}
Q.~Sun, Z.~Liu, C.~Ma, Z.~Ding, F.~Xu, Z.~Yin, H.~Zhao, Z.~Wu, K.~Cheng,
  Z.~Liu, J.~Wang, Q.~Li, X.~Tang, T.~Xie, X.~Feng, X.~Li, B.~Kao, W.~Wang,
  B.~Qi, L.~Kong, and Z.~Wu.
\newblock Scienceboard: Evaluating multimodal autonomous agents in realistic
  scientific workflows.
\newblock In \emph{The Fourteenth International Conference on Learning
  Representations}, 2026{\natexlab{c}}.
\newblock URL \url{https://openreview.net/forum?id=bJvwJahJeF}.

\bibitem[Wan et~al.(2026)Wan, Zhang, Wang, and Zhu]{wan2026seekjudge}
Y.~Wan, Z.~Zhang, J.~Wang, and L.~Zhu.
\newblock Seekjudge: A practical reward framework for reinforcement learning in
  computer-use agents.
\newblock \emph{arXiv preprint arXiv:2607.23263}, 2026.

\bibitem[Xie et~al.(2024)Xie, Zhang, Chen, Li, Zhao, Cao, Hua, Cheng, Shin,
  Lei, Liu, Xu, Zhou, Savarese, Xiong, Zhong, and Yu]{OSWorld}
T.~Xie, D.~Zhang, J.~Chen, X.~Li, S.~Zhao, R.~Cao, T.~J. Hua, Z.~Cheng,
  D.~Shin, F.~Lei, Y.~Liu, Y.~Xu, S.~Zhou, S.~Savarese, C.~Xiong, V.~Zhong, and
  T.~Yu.
\newblock {OSW}orld: Benchmarking multimodal agents for open-ended tasks in
  real computer environments.
\newblock In \emph{The Thirty-eight Conference on Neural Information Processing
  Systems Datasets and Benchmarks Track}, 2024.
\newblock URL \url{https://openreview.net/forum?id=tN61DTr4Ed}.

\bibitem[Xie et~al.(2025)Xie, Yuan, Zhang, Xiong, Shen, Zhou, Wang, Chen, Deng,
  Chen, Wang, Wu, Chen, Wang, Lu, Hu, and Yu]{osworldverified}
T.~Xie, M.~Yuan, D.~Zhang, X.~Xiong, Z.~Shen, Z.~Zhou, X.~Wang, Y.~Chen,
  J.~Deng, J.~Chen, B.~Wang, H.~Wu, J.~Chen, J.~Wang, D.~Lu, H.~Hu, and T.~Yu.
\newblock Introducing osworld-verified.
\newblock \emph{xlang.ai}, July 2025.
\newblock URL \url{https://xlang.ai/blog/osworld-verified}.

\bibitem[Xue et~al.(2026{\natexlab{a}})Xue, Liao, Shi, Wang, Zhang, Song, Su,
  and Sun]{xue2026acurl}
T.~Xue, Z.~Liao, T.~Shi, Z.~Wang, K.~Zhang, D.~Song, Y.~Su, and H.~Sun.
\newblock Autonomous continual learning for environment adaptation of
  computer-use agents, 2026{\natexlab{a}}.
\newblock URL \url{https://arxiv.org/abs/2602.10356}.

\bibitem[Xue et~al.(2026{\natexlab{b}})Xue, Peng, Huang, Guo, Han, Wang, Wang,
  Zhang, Yang, Zhao, Ding, Ma, Xie, Pei, Cai, and
  Qiu]{xue2026evocuaevolvingcomputeruse}
T.~Xue, C.~Peng, M.~Huang, L.~Guo, T.~Han, H.~Wang, J.~Wang, X.~Zhang, X.~Yang,
  D.~Zhao, J.~Ding, X.~Ma, Y.~Xie, P.~Pei, X.~Cai, and X.~Qiu.
\newblock Evocua: Evolving computer use agents via learning from scalable
  synthetic experience, 2026{\natexlab{b}}.
\newblock URL \url{https://arxiv.org/abs/2601.15876}.

\bibitem[Zhou et~al.(2024)Zhou, Xu, Zhu, Zhou, Lo, Sridhar, Cheng, Ou, Bisk,
  Fried, Alon, and Neubig]{webarena}
S.~Zhou, F.~F. Xu, H.~Zhu, X.~Zhou, R.~Lo, A.~Sridhar, X.~Cheng, T.~Ou,
  Y.~Bisk, D.~Fried, U.~Alon, and G.~Neubig.
\newblock Webarena: A realistic web environment for building autonomous agents.
\newblock In \emph{The Twelfth International Conference on Learning
  Representations}, 2024.
\newblock URL \url{https://openreview.net/forum?id=oKn9c6ytLx}.

\bibitem[Zhuge et~al.(2025)Zhuge, Zhao, Ashley, Wang, Khizbullin, Xiong, Liu,
  Chang, Krishnamoorthi, Tian, Shi, Chandra, and
  Schmidhuber]{zhuge2025agentasajudge}
M.~Zhuge, C.~Zhao, D.~R. Ashley, W.~Wang, D.~Khizbullin, Y.~Xiong, Z.~Liu,
  E.~Chang, R.~Krishnamoorthi, Y.~Tian, Y.~Shi, V.~Chandra, and J.~Schmidhuber.
\newblock Agent-as-a-judge: Evaluate agents with agents.
\newblock In \emph{Forty-second International Conference on Machine Learning},
  2025.
\newblock URL \url{https://openreview.net/forum?id=Nn9POI9Ekt}.

\end{thebibliography}

\clearpage
\appendix
\numberwithin{table}{section}
\numberwithin{figure}{section}
\renewcommand{\theHtable}{appendix.\thesection.\arabic{table}}
\renewcommand{\theHfigure}{appendix.\thesection.\arabic{figure}}
\section*{Appendix Contents}
\startcontents[appendices]
\printcontents[appendices]{}{1}{\setcounter{tocdepth}{2}}
\clearpage

\section{Implementation Details}

\label{app:env}

\FloatBarrier
\subsection{Verification Settings}
\label{app:impl-settings}

Table~\ref{tab:corr} lists what each setting determines at each stage of the
procedure.

\begin{table}[!ht]
\centering
\small
\caption{What each setting determines in the shared procedure.}
\label{tab:corr}
\setlength{\tabcolsep}{5pt}
\renewcommand{\arraystretch}{1.2}
\begin{tabularx}{\linewidth}{@{}
  >{\raggedright\arraybackslash}p{2.05cm}
  >{\raggedright\arraybackslash}p{2.5cm}
  >{\raggedright\arraybackslash}X
  >{\raggedright\arraybackslash}X@{}}
\toprule
\textbf{Stage} & \textbf{Aspect} & \textbf{Online environment} & \textbf{Recorded trajectory} \\
\midrule
\textbf{Inspect context} & application context & packages, directories, stores, and exposed schemas & recorded steps, frames, and structured observations \\
\addlinespace[5pt]
\textbf{Synthesize rubric} & grounded task requirements & instruction and observed machine context & instruction and observed recording context \\
\addlinespace[5pt]
\textbf{Explore} & evidence gathering & adaptive GUI reads and code checks, including synthesized checkers & up to $K$ audit paths, seeking support and refutations \\
\addlinespace[2pt]
 & attested types & environment read, UI dump, harness artifact & post-effect frame, precise read, computation, structured state \\
\addlinespace[2pt]
 & source attributes & persistence, persistent or volatile & frame timing, before or after the relevant action \\
\addlinespace[5pt]
\textbf{Arbitrate} & requirement confirmation & one persistent attested confirmation & attested support from every completed path \\
\addlinespace[2pt]
 & unresolved task & abstention, counted as \textsc{fail} in binary evaluation & initial-judgment fallback \\
\bottomrule
\end{tabularx}
\end{table}

\FloatBarrier
\Needspace{0.55\textheight}
\subsection{Action Space and Evidence Types}
\label{app:impl-actions}

Table~\ref{tab:actions} lists the verbs by state source and modality,
together with the evidence type each one stamps.

\begin{table}[!ht]
\centering
\small
\caption{The action space by state source and modality, with the evidence
type each verb stamps.}
\label{tab:actions}
\setlength{\tabcolsep}{4pt}
\renewcommand{\arraystretch}{1.2}
\begin{tabularx}{\linewidth}{@{}
  >{\raggedright\arraybackslash}p{2.45cm}
  >{\raggedright\arraybackslash}p{1.9cm}
  >{\hsize=1.12\hsize\linewidth=\hsize\raggedright\arraybackslash}X
  >{\hsize=0.88\hsize\linewidth=\hsize\raggedright\arraybackslash}X
  >{\raggedright\arraybackslash}p{1.45cm}@{}}
\toprule
\textbf{State Source} & \textbf{Modality} & \textbf{Verbs} & \textbf{Evidence Type} & \textbf{Class} \\
\midrule
\textbf{actions and reasoning} & text & \texttt{find\_steps}, \texttt{inspect\_action} & agent claim & asserted \\
 & code & \texttt{compute} & sandboxed computation & attested \\
\addlinespace[5pt]
\textbf{screen frames} & GUI & \texttt{overview}, \texttt{get\_frame}, \texttt{zoom}, \texttt{ocr} & post-effect frame; precise read & attested \\
 & & & pre-action frame & asserted \\
\addlinespace[5pt]
\textbf{\texttt{a11ytrees}} & structured & \texttt{read\_a11y}, \texttt{search\_a11y} & structured state & attested \\
\midrule

\textbf{machine} & code & \texttt{file\_exists}, \texttt{find}, \texttt{list\_dir}, \texttt{file\_stat}, \texttt{read\_file}, \texttt{sqlite\_query}, \texttt{content\_query} & environment read (persistent) & attested \\
\addlinespace[3pt]
 & code & \texttt{dumpsys}, \texttt{current\_activity}, \texttt{settings\_get}, \texttt{getprop}, \texttt{pm\_list} & environment read (persistent or volatile) & attested \\
\addlinespace[3pt]
 & GUI, structured & \texttt{screenshot}, \texttt{ui\_dump} & screenshot; UI dump (volatile) & attested \\
\addlinespace[3pt]
 & text & \texttt{submitted\_answer}, \texttt{clipboard\_read} & harness artifact & attested \\
\bottomrule
\end{tabularx}
\end{table}

The getter inventory has
seventeen verbs, listed in Table~\ref{tab:getter-persistence} with their persistence tags. The image-reading
verbs are enabled in the vision fallback configuration,
\texttt{clipboard\_read} runs last because it disturbs the screen, and a
miss from \texttt{find} counts only as weak evidence.

\begin{table}[!ht]
\centering
\footnotesize
\caption{Persistence tags of the online getter inventory.}
\label{tab:getter-persistence}
\begin{tabular}{@{}lp{10.6cm}@{}}
\toprule
persistent & \texttt{settings\_get}, \texttt{getprop}, \texttt{file\_exists}, \texttt{file\_stat}, \texttt{read\_file}, \texttt{list\_dir}, \texttt{find}, \texttt{sqlite\_query}, \texttt{content\_query}, \texttt{pm\_list}, \texttt{view\_file\_image} \\
volatile & \texttt{dumpsys}, \texttt{ui\_dump}, \texttt{current\_activity}, \texttt{clipboard\_read}, \texttt{submitted\_answer}, \texttt{screencap} (the screenshot verb of Table~\ref{tab:actions}) \\
\bottomrule
\end{tabular}
\end{table}

\paragraph{Extension Contract.}

A new source enters as exactly (i) its read verbs, (ii) its evidence types
with their attested or asserted classification, and (iii) an initial view
for the cold-start judgment; every later stage consumes types, so nothing
downstream changes. The accessibility channel entered this way, and the
machine channel repeated the pattern with its own inventory.

\FloatBarrier
\subsection{Evidence Aggregation}
\label{app:impl-adjudication}

On recorded trajectories, up to $K$ audit paths gather support through
forward inspection and seek refutations through backward inspection.
The default $K=3$ permits early stopping after two paths
agree on every hard requirement. A requirement is \textsc{met} only through attested
support from every completed path and \textsc{unmet} only with corroborating
refutations from separate probes. If the evidence does not settle the task,
the initial judgment supplies the fallback verdict without adding attested
support to \textsc{unknown} requirements.

In online environments, initial inspection maps the device's packages, stores,
and schemas before rubric synthesis. The observed context informs the rubric
and subsequent checker construction; each requirement receives checkers on
more than one surface where possible; predicates are evaluated by a restricted evaluator
or, for semantic comparisons, by a judge call; a requirement is \textsc{met} by one
persistent attested confirmation and \textsc{unmet} by two corroborating refutations
(the \texttt{unmet\_min} threshold, which trades decided episodes for
precision when raised). An unresolved task produces an abstention, which
binary evaluation counts as \textsc{fail}. Across passes, the combination rule takes the decided verdict and
abstains when two decided verdicts disagree.

\FloatBarrier
\subsection{Read-Only Enforcement}
\label{app:impl-readonly}

On recordings the read-only property is structural, since every verb is a
pure function of files on disk. On a machine the verbs come from a fixed
allowlist, arguments that could smuggle shell control are rejected, and a
digest of the plan's persistent targets is compared before and after
execution.

\FloatBarrier
\subsection{Online Environment Adapters}
\label{app:impl-environments}

\paragraph{AndroidWorld.}

In online environments, VeriCUA runs against AndroidWorld on an Android emulator. The
harness has root access over \texttt{adb}, so every store the benchmark's own
evaluator can read is readable by the verifier. Each episode is executed by
the benchmark's agent scaffold; when it ends, the harness saves the device
state as a snapshot and verification runs against that snapshot, so any
verification pass can be replayed on the identical end state. Before the
episode starts, the harness records a pre-state anchor of the application
storage, against which requirements about a change are compared. The agent's submitted answer is extracted by the
harness and exposed through the \texttt{submitted\_answer} getter as the
artifact under test.

The emulator is a headless Pixel 6 image at API level 33 (Android 13) with
root access over \texttt{adb}. Snapshots are taken through the emulator
console once the official evaluator has run.

\paragraph{OSWorld and ScienceBoard.}

OSWorld runs in a virtual machine and ScienceBoard in its own scientific
desktop environment. Both are verified through the same allowlisted read-only
getters over the machine, and ScienceBoard's evaluator reads the state of the
scientific applications inside the VM. Both use the virtual machine images
that the benchmarks ship, without modification.

\FloatBarrier
\subsection{Recorded Observation Processing}
\label{app:impl-recorded}

A recorded trajectory enters as the agent's actions and reasoning, its
screen frames, and, when the platform recorded them, its \texttt{a11ytrees}.
Frame semantics, whether frame $i$ captures the state before or after action
$i$, is declared per dataset and determines the post-effect and pre-action
evidence types. The \texttt{a11ytrees} are condensed before they are read.
OSReward trajectories are pure GUI with pre-action frame semantics; WebArena
trajectories carry \texttt{a11ytrees}.

The condenser writes one line per node of the browser's accessibility dump,
indented by depth, with the node id, role, name, and ten retained properties
such as value, checked, focused, and url. Ignored nodes and nameless nodes
without properties are dropped, and names are cut to 110 characters. A read
returns at most 150 lines with a note on how many nodes were left out, a
query returns at most 40 matching lines, and a search across steps returns
the first match in each of up to 12 steps. Each trajectory enters as one
record that lists, per step, the agent's thought and action, the page URL,
and the paths of the screenshot and the tree.

\FloatBarrier

\Needspace{0.4\textheight}
\section{Experimental Setup and Evaluation Protocol}

\label{app:exp}

\FloatBarrier
\subsection{Evaluation Sets}
\label{app:exp-data}

Table~\ref{tab:datasets} summarizes the evaluation sets.

\begin{table}[!ht]
\centering
\footnotesize
\caption{Evaluation sets.}
\label{tab:datasets}
\setlength{\tabcolsep}{5pt}
\renewcommand{\arraystretch}{1.2}
\begin{tabular}{@{}lll>{\raggedright\arraybackslash}p{4.6cm}@{}}
\toprule
\textbf{Set} & \textbf{Size} & \textbf{Gold} & \textbf{Notes} \\
\midrule
OSReward-Hard & 284 & human labels & hard subset of OSReward \\
OSReward full & 1019 & human labels & \\

WebArena & 715 & official checkers &
  239 / 235 / 241 rollouts by Claude Sonnet 4.6, Gemini 3.1 Pro, Gemini 3 Flash \\

AndroidWorld  & 116 & official evaluator & M3A Agents \\
OSWorld, online & 361 & official evaluator & one rollout per task; actor success 44.3\% \\
ScienceBoard, online & 169 & official evaluator & actor success 23.7\% \\
\bottomrule
\end{tabular}
\end{table}

\paragraph{Multi-Pass Suite.}

The multi-pass suite keeps every snapshot on
which the agent claimed completion but gold says \textsc{fail}, on which the
configurations disagree, or on which any pass abstained.

\FloatBarrier
\subsection{Baseline Implementations}
\label{app:exp-baselines}

\paragraph{Offline Verification.}

TRM scores the trajectory from 1 to 5 in a single call from the instruction,
every thought and action, and the last three frames; CUAJudge checks up to
six task-derived key points against the first and last of up to six frames
in a single call; OS-Themis verifies up to four ordered milestones over up to
eight frames and reviews them in a second call.

The three offline judges run under one harness at temperature 0,
with images downsampled to a long side of 1280 pixels, the first and last
frames pinned, no self-consistency voting, and a parse failure mapped to
\textsc{fail}. TRM thresholds its 1--5 score at 4 and is the only judge that sees the
agent's thoughts. CUAJudge collapses the original four-stage pipeline into
one call, dropping the dedicated frame-selection and init-versus-final
modules. OS-Themis collapses the multi-agent critic to two calls with the
budget cut to four milestones and eight frames against the original six and
twelve; Table~\ref{tab:themis} shows that restoring the original budget
changes little.

\paragraph{Online Environments.}

The online VLM judge answers
in one word from the instruction and the final screenshot and never
abstains; the auto-generated verifier writes one evaluator script per task
with full device access and runs it once.

The auto-generated verifier receives the instruction
with its concrete parameters, the same generic device context our synthesis
receives, unrestricted host-side \texttt{adb} and Python, and the agent's
submitted answer on answer-scored tasks. It does not receive the trajectory,
the results of context inspection, or the pre-state anchor, and it runs last on each end state.

\paragraph{Majority Voting.}

Majority voting combines the same three verifier draws by vote with
abstentions folded to \textsc{fail}.

\FloatBarrier
\subsection{Verifier Configuration}
\label{app:exp-config}

\paragraph{Offline Verification.}

The default runs $K = 3$ paths and stops after two when they agree on the
status of every hard requirement, which happens on about two fifths of the
trajectories. A path spends up to four tool calls gathering support and then
at least four, or two per hard requirement, hunting refutations. Each call
returns one observation, frames are downscaled to at most 1288 pixels on the
long side, and only the last eight images stay in context. 

The rubric comes
from three candidate decompositions consolidated in a fourth call. The
sealed initial judgment reads the last five frames together with the
thought and action log and is reused only as the fallback.

\paragraph{Online Environments.}

Rubric synthesis draws two candidate rubrics at temperature 0.7 and
consolidates them in a third call at temperature 0; checker synthesis makes
two parallel calls at temperature 0.7 and asks for two checkers per hard
requirement on different surfaces where the state allows it; context inspection
uses one call over at most three packages and six databases, followed by one
schema probe; a semantic predicate costs one judge call at temperature 0
over at most three images. 

No call carries a token cap, and each has a
180-second timeout with up to six retries.

\FloatBarrier
\subsection{Evaluation Metrics}
\label{app:exp-metrics}

The binary evaluation treats \textsc{success} as the positive class.
Abstentions are counted as \textsc{fail} for agreement, precision, recall,
F1, and the false positive rate. For verifiers that abstain, we also report
the decision rate and the accuracy on decided episodes.

\FloatBarrier

\FloatBarrier
\section{Additional Results and Analysis}
\label{app:results}

\FloatBarrier
\subsection{Action Space and Evidence Rules}
\label{app:ana-action}

\paragraph{Action-Space Ablation.}
The values behind Figure~\ref{fig:action} are in Table~\ref{tab:actionspace}.

\begin{table}[!ht]
\centering
\small
\caption{Action-space ablation in the online AndroidWorld environment.}
\label{tab:actionspace}
\setlength{\tabcolsep}{5pt}
\renewcommand{\arraystretch}{1.15}
\begin{tabular}{@{}lccccccc@{}}
\toprule
\textbf{Action Space} & \textbf{Prec} & \textbf{Rec} & \textbf{F1} & \textbf{Acc} &
\textbf{False Positives} & \textbf{Decided} & \textbf{Decided Acc} \\
\midrule
GUI only & 86.4 & 57.6 & 69.1 & 77.0 & 3/41 & 45\% & 87.9\% \\
\textbf{Hybrid (default)} & \textbf{90.6} & \textbf{87.9} & \textbf{89.2} & \textbf{90.5} & 3/41 & 82\% & \textbf{95.1\%} \\
\bottomrule
\end{tabular}
\end{table}

\begin{figure}[!ht]
\centering
\includegraphics[width=\textwidth]{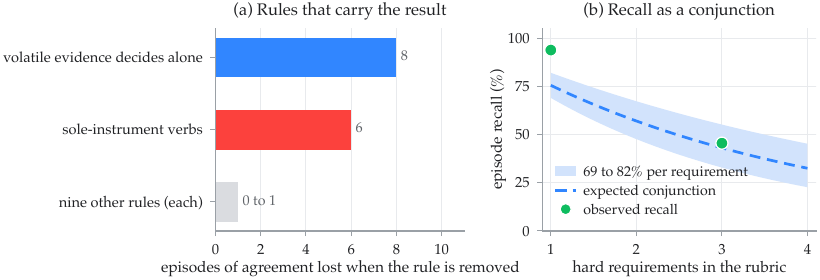}
\caption{Evidence rules and recall on the first AndroidWorld suite.}
\label{fig:rules}
\end{figure}

\paragraph{Evidence Rules And Recall.}
Figure~\ref{fig:rules} shows which arbiter rules carry the result on the
first AndroidWorld suite and how episode recall falls with the number of
hard requirements.

\FloatBarrier
\Needspace{0.35\textheight}
\subsection{Observation Inputs and Rollout Agents}
\label{app:ana-observation}

\paragraph{VeriCUA By Observation Input.}
Table~\ref{tab:obs_input} gives VeriCUA's accuracy, decision rate, share of
verdicts resting on structured evidence, and cost under each observation
input on WebArena.

\begin{table}[!ht]
\centering
\small
\caption{VeriCUA by observation input on WebArena.}
\label{tab:obs_input}
\setlength{\tabcolsep}{6pt}
\renewcommand{\arraystretch}{1.15}
\begin{tabular}{@{}lccccc@{}}
\toprule
\textbf{Input} & \textbf{Acc} & \textbf{BalAcc} & \textbf{Decided} & \textbf{Structured Evidence} & \textbf{Tokens} \\
\midrule
Screenshots & 80.0 & 80.0 & 33\% & 0\% & 234K \\
Screenshots + \texttt{a11ytrees} & 80.0 & 80.0 & 75\% & 73\% & 252K \\
\texttt{a11ytrees} & 76.7 & 76.7 & 85\% & 93\% & 178K \\
\bottomrule
\end{tabular}
\end{table}

\paragraph{WebArena Per Rollout Agent.}
The WebArena results combine trajectories collected from three rollout
agents. Table~\ref{tab:webarena} gives the results for each agent and for
the combined set.

\begin{table}[!ht]
\centering
\small
\caption{WebArena results by rollout agent and overall.}
\label{tab:webarena}
\setlength{\tabcolsep}{4pt}
\renewcommand{\arraystretch}{1.15}
\begin{tabular}{@{}lcccccccc@{}}
\toprule
 & \multicolumn{6}{c}{\textbf{Rollout Agent}} & \multicolumn{2}{c}{\multirow{2}{*}{\textbf{Overall}}} \\
\cmidrule(lr){2-7}
 & \multicolumn{2}{c}{Claude Sonnet 4.6} & \multicolumn{2}{c}{Gemini 3.1 Pro} &
\multicolumn{2}{c}{Gemini 3 Flash} & \multicolumn{2}{c}{} \\
\cmidrule(lr){2-3} \cmidrule(lr){4-5} \cmidrule(lr){6-7}
 & Acc & BalAcc & Acc & BalAcc & Acc & BalAcc & Acc & BalAcc \\
\midrule
\textbf{VeriCUA} (GUI + \texttt{a11ytree}) & \textbf{79.9} & \textbf{80.3} & \textbf{84.3} &
\textbf{85.3} & \textbf{83.8} & \textbf{83.9} & \textbf{82.7} & \textbf{83.1} \\
CUAJudge & 79.1 & 79.8 & 82.1 & 83.4 & 82.6 & 81.9 & 81.3 & 81.8 \\
TRM & 76.2 & 77.0 & 79.1 & 80.8 & 83.4 & 81.0 & 79.6 & 79.8 \\
OS-Themis & 70.3 & 69.8 & 80.0 & 79.4 & 75.9 & 74.1 & 75.4 & 74.3 \\
\bottomrule
\end{tabular}
\end{table}

\paragraph{Tree-Fed Baselines On WebArena.}
The judges are re-run with \texttt{a11ytrees} alone as their input, in place of the
GUI and \texttt{a11ytree} input of Table~\ref{tab:webarena}, per rollout agent and
overall (Table~\ref{tab:webarena_tree}). VeriCUA retains its GUI and
\texttt{a11ytree} input in this comparison.

\begin{table}[!ht]
\centering
\small
\caption{Accuracy and balanced accuracy on WebArena with the judges reading \texttt{a11ytrees}.}
\label{tab:webarena_tree}
\setlength{\tabcolsep}{4pt}
\renewcommand{\arraystretch}{1.15}
\begin{tabular}{@{}lcccccccc@{}}
\toprule
 & \multicolumn{2}{c}{Claude Sonnet 4.6} & \multicolumn{2}{c}{Gemini 3.1 Pro} &
\multicolumn{2}{c}{Gemini 3 Flash} & \multicolumn{2}{c}{Overall} \\
\cmidrule(lr){2-3} \cmidrule(lr){4-5} \cmidrule(lr){6-7} \cmidrule(lr){8-9}
 & Acc & BalAcc & Acc & BalAcc & Acc & BalAcc & Acc & BalAcc \\
\midrule
\textbf{VeriCUA} (GUI + \texttt{a11ytree}) & \textbf{79.9} & \textbf{80.3} & \textbf{84.3} & \textbf{85.3} & \textbf{83.8} & \textbf{83.9} & \textbf{82.7} & \textbf{83.1} \\
CUAJudge, tree-fed & 77.8 & 77.9 & 83.0 & 84.3 & 82.6 & 80.0 & 81.1 & 80.8 \\
TRM, tree-fed & 74.5 & 74.9 & 80.4 & 81.4 & 80.9 & 77.4 & 78.6 & 78.1 \\
OS-Themis, tree-fed & 74.5 & 73.5 & 79.6 & 78.7 & 78.0 & 74.9 & 77.3 & 75.7 \\
\bottomrule
\end{tabular}
\end{table}

\FloatBarrier
\Needspace{0.5\textheight}
\subsection{Test-Time Scaling and Budget}
\label{app:ana-tts}

\paragraph{Multi-Pass Combination In The Online Environment.}
The multi-pass AndroidWorld suite compares repeated runs of the default
code and structured-state configuration with a run that enables visual
inspection, called the vision fallback. These verifier runs and the
auto-generated baseline share one restored snapshot per episode.
Table~\ref{tab:multipass} reports individual and combined results. Adding
visual inspection improves agreement and decision rate over repeated
default runs, including at a matched pass count. The combination rule also
produces fewer incorrect \textsc{fail} verdicts than majority voting.

\begin{table}[!ht]
\centering
\footnotesize
\caption{Multi-pass combination in the online environment.}
\label{tab:multipass}
\setlength{\tabcolsep}{5pt}
\renewcommand{\arraystretch}{1.15}
\begin{tabular}{@{}lccccc@{}}
\toprule
\textbf{Pass} & \textbf{Agreement} & \makecell{\textbf{Decision}\\\textbf{Rate}} &
\makecell{\textbf{Decided}\\\textbf{Acc}} & \makecell{\textbf{False}\\\textbf{Positives}} &
\makecell{\textbf{Wrong}\\\textbf{\textsc{fail}s}} \\
\midrule
draw 1 & 75.7 & 72.2 & 91.6 & 3 & 4 \\
draw 2 & 79.1 & 77.4 & 92.1 & 3 & 4 \\
draw 3 & 77.4 & 67.0 & 94.8 & 1 & 3 \\
draw with vision fallback & 76.5 & 73.0 & 90.5 & 3 & 5 \\
three draws combined & 80.9 & 78.3 & 95.6 & 2 & 2 \\
\textbf{three draws + vision fallback} & \textbf{87.0} & \textbf{87.0} &
  \textbf{96.0} & 2 & 2 \\
draws 1, 2 + vision (equal cost) & 84.3 & 85.2 & 94.9 & 3 & 2 \\
\midrule
three draws, majority vote & 80.0 & 100 & 80.0 & 3 & 20 \\
auto-generated verifier & 78.3 & 100 & 78.3 & 4 & 21 \\
\bottomrule
\end{tabular}
\end{table}

\paragraph{Baseline Budget.}
Table~\ref{tab:themis} reports the OS-Themis budget ablation.

\begin{table}[!ht]
\centering
\small
\caption{OS-Themis budget ablation, mean balanced accuracy.}
\label{tab:themis}
\setlength{\tabcolsep}{8pt}
\renewcommand{\arraystretch}{1.15}
\begin{tabular}{@{}lc@{}}
\toprule
\textbf{Configuration} & \textbf{BalAcc} \\
\midrule
OS-Themis, 4 milestones / 8 frames (used in the main text) & 74.2 \\
OS-Themis, 6 milestones / 12 frames (original budget) & 75.9 \\
CUAJudge & 76.8 \\
\bottomrule
\end{tabular}
\end{table}

\FloatBarrier
\Needspace{0.35\textheight}
\subsection{Backbone Comparisons}
\label{app:ana-backbone}

\paragraph{Single-Call Judges On Other Backbones.}
Results for the single-call judge of OSReward on several backbones appear
alongside VeriCUA on \texttt{gemini-3-flash-preview} in
Table~\ref{tab:osreward_judges}.

\begin{table}[!ht]
\centering
\small
\caption{The OSReward single-call judge by backbone, with VeriCUA for reference.}
\label{tab:osreward_judges}
\setlength{\tabcolsep}{7pt}
\renewcommand{\arraystretch}{1.15}
\begin{tabular}{@{}lcccc@{}}
\toprule
 & \multicolumn{2}{c}{\textbf{OSReward-Hard}} & \multicolumn{2}{c}{\textbf{OSReward full}} \\
\cmidrule(lr){2-3} \cmidrule(lr){4-5}
\textbf{Judge} & Acc & BalAcc & Acc & BalAcc \\
\midrule
\textbf{VeriCUA} (Gemini 3 Flash) & \textbf{70.4} & 68.3 & 87.2 & 86.3 \\
\midrule
Gemini 3 Flash & 57.0 & 65.2 & 87.0 & 88.2 \\
Gemini 3.1 Pro & 61.6 & 61.6 & 87.9 & 88.2 \\
Claude Sonnet 4.6 & 59.2 & 68.1 & 87.7 & 88.9 \\
Claude Opus 4.6 & 67.3 & 68.6 & 89.5 & 90.2 \\
Claude Opus 4.8 & 69.7 & \textbf{69.7} & \textbf{89.7} & 90.0 \\
GPT-5.5 & 67.3 & 67.0 & 89.5 & 89.8 \\
\bottomrule
\end{tabular}
\end{table}

\paragraph{Backbone Study In The Online Environment.}
Table~\ref{tab:backbone_online} repeats the flash-to-pro comparison of
Table~\ref{tab:backbone} in the online AndroidWorld environment.

\begin{table}[!ht]
\centering
\small
\caption{Accuracy and balanced accuracy in the online AndroidWorld environment
with each framework on the flash and pro tiers.}
\label{tab:backbone_online}
\setlength{\tabcolsep}{6pt}
\renewcommand{\arraystretch}{1.15}
\begin{tabular}{@{}lcccccc@{}}
\toprule
\multirow{2}{*}{\textbf{Judge}} & \multicolumn{3}{c}{\textbf{Acc}} & \multicolumn{3}{c}{\textbf{BalAcc}} \\
\cmidrule(lr){2-4} \cmidrule(lr){5-7}
 & \textbf{Flash} & \textbf{Pro} & $\Delta$ & \textbf{Flash} & \textbf{Pro} & $\Delta$ \\
\midrule
VLM judge & 78.4 & 79.7 & +1.3 & 77.5 & 78.5 & +1.0 \\
Auto-generated verifier & 82.4 & 85.1 & +2.7 & 82.7 & 85.4 & +2.7 \\
\textbf{VeriCUA} & \textbf{90.5} & \textbf{91.9} & +1.4 & \textbf{90.3} & \textbf{91.8} & +1.5 \\
\bottomrule
\end{tabular}
\end{table}

\FloatBarrier
\Needspace{0.4\textheight}
\subsection{Computational Cost}
\label{app:ana-cost}

Token usage per trajectory or episode is reported in Table~\ref{tab:cost}.

\begin{table}[!ht]
\centering
\small
\caption{Cost per trajectory or episode.}
\label{tab:cost}
\setlength{\tabcolsep}{7pt}
\renewcommand{\arraystretch}{1.15}
\begin{tabular}{@{}llr@{}}
\toprule
\textbf{Setting} & \textbf{Method} & \textbf{Tokens} \\
\midrule
offline, WebArena (tree-fed, flash) & three judges together (4 calls) & 43K--61K \\
offline, WebArena (GUI, pro) & three judges together (4 calls) & 25K \\
offline, OSReward (GUI) & three judges together (4 calls) & 33K--35K \\
offline, WebArena & VeriCUA & 150K--208K \\
offline, OSReward & VeriCUA & 345K--378K \\
\midrule
online, first suite & auto-generated verifier & 16.2K (60 s) \\
online, first suite & VeriCUA & 39.2K (97 s) \\
online, multi-pass suite & auto-generated verifier & 21K \\
online, multi-pass suite & VeriCUA, per pass & 43K--51K \\
\bottomrule
\end{tabular}
\end{table}

On AndroidWorld, a snapshot load restores the end state in about fifteen
seconds. One verifier pass
takes about 100 seconds per episode, of which checker synthesis takes about
half and predicate evaluation about a third, with getter execution and
context inspection taking under ten seconds together.

\FloatBarrier
\subsection{Confident Failures and Gold-Label Audit}
\label{app:ana-audit}

\paragraph{Confident Failure Verdicts.}
For offline verification, we start with all initial \textsc{fail} judgments and retain
those for which exploration also yields a decided \textsc{fail}. Initial judgments
use the last five available frames. Evidence aggregation uses
three paths, a \textsc{met} threshold of 0.67, and an \textsc{unmet} fraction of 0.4.

Failure precision is the fraction of predicted \textsc{fail} verdicts that match
the reference labels. Retention is the selected subset size divided by
the number of initial \textsc{fail} judgments. Figure~\ref{fig:failure-precision}
connects these two selections within each cohort; Table~\ref{tab:tier}
provides their counts. WebArena uses its official reference labels, and
OSReward uses human labels.
OSReward-Hard is included in OSReward full.

\begin{table}[!ht]
\centering
\small
\caption{Initial failure judgments and the subset retained after exploration.}
\label{tab:tier}
\setlength{\tabcolsep}{4pt}
\renewcommand{\arraystretch}{1.15}
\begin{tabular}{@{}lcccccc@{}}
\toprule
\multirow{2}{*}{\textbf{Dataset}} & \multirow{2}{*}{\textbf{Total $N$}} &
\multicolumn{2}{c}{\textbf{Initial \textsc{fail}}} &
\multicolumn{2}{c}{\textbf{Selected \textsc{fail}}} &
\multirow{2}{*}{\textbf{Retained (\%)}} \\
\cmidrule(lr){3-4}\cmidrule(lr){5-6}
 & & Correct / $n$ & Prec. (\%) & Correct / $n$ & Prec. (\%) & \\
\midrule
WebArena & 715 & 381/431 & 88.4 & 195/212 & 92.0 & 49.2 \\
OSReward full & 1019 & 524/625 & 83.8 & 166/179 & 92.7 & 28.6 \\
OSReward-Hard & 284 & 148/185 & 80.0 & 48/54 & 88.9 & 29.2 \\
\bottomrule
\end{tabular}
\end{table}

\paragraph{Gold-Label Audit.}
The gold-label audit uses 814 recorded trajectories from AndroidWorld,
OSWorld, and WebArena. A trajectory is flagged when a decided VeriCUA
verdict contradicts the benchmark's gold label, which happens on 81 of them.
A single-blind adjudicator, a model from
a different family than the verifier's backbone, then re-judges each flagged
trajectory from scratch. It sees only the instruction, the agent's thought
and action log, and up to ten evenly spaced frames including the last one,
never the verifier's verdict, the gold label, or the verifier's reasoning,
and it may answer \textsc{success}, \textsc{fail}, or \textsc{unsure}.

A flag counts as confirmed when
the adjudicator sides with VeriCUA against gold. Of the 81 flags, 51 are
confirmed, 20 refuted, and 10 left \textsc{unsure}, with \textsc{unsure} kept in the
denominator. On OSWorld, the adjudicator confirms 37 of 57 flags. A second
adjudicator, run under the same blind protocol, confirms 34 of the 51 flags
it decides, excluding its \textsc{unsure} cases from that denominator.
In the online OSWorld environment,
about 18 of VeriCUA's 28 false positives are episodes that the official
evaluator marks \textsc{fail} in error.

\FloatBarrier

\FloatBarrier
\begingroup
\fontsize{10}{11.5}\selectfont
\setlength{\parindent}{0pt}
\setlength{\parskip}{0.46em}
\setlength{\textfloatsep}{0.55em}
\setlength{\intextsep}{0.5em}
\Needspace{0.66\textheight}
\section{Case Studies}
\label{app:cases}

Each card shows the instruction and the frames or machine reads the
verifier cited. The summary below it gives the verdict against gold and the
baselines, the rubric with each requirement's final status, and the typed
evidence the arbiter used.

\FloatBarrier
\Needspace{0.6\textheight}
\subsection{Online Environments}
\label{app:cases-online}

The verifier reads the live machine after the episode ends.

\casecard{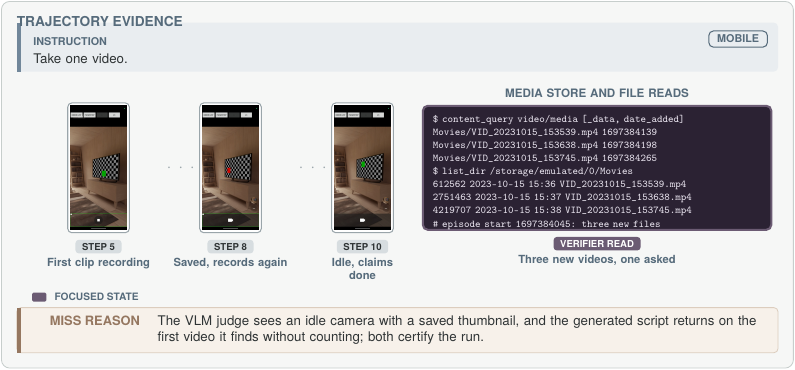}
{fig:case-camera}
{\textbf{Gold}\enspace\textsc{fail}\quad\textbf{Agent}\enspace claimed complete\hfill
 \textbf{VLM judge}\enspace\textsc{success}\quad\textbf{Generated verifier}\enspace\textsc{success}\quad\textbf{VeriCUA}\enspace\textsc{fail}}
{\rubricline{\runmet}{Exactly one new video since the episode began.}
 \rubricline{\runknown}{The new video sits in a standard media directory.}}
{\evidenceline{\texttt{content\_query} on the video media store, persistent:
  three rows added after the start.}
 \evidenceline{\texttt{list\_dir} and \texttt{find} over \texttt{Movies},
  persistent: the same three files, written one minute apart.}
 \evidenceline{Three of the four verifier passes decide \textsc{fail} on
  this evidence; the fourth abstains.}}
{A false completion claim refuted from the media store. The agent recorded
three clips for one requested video, and both baselines certified the run.}

\casecard{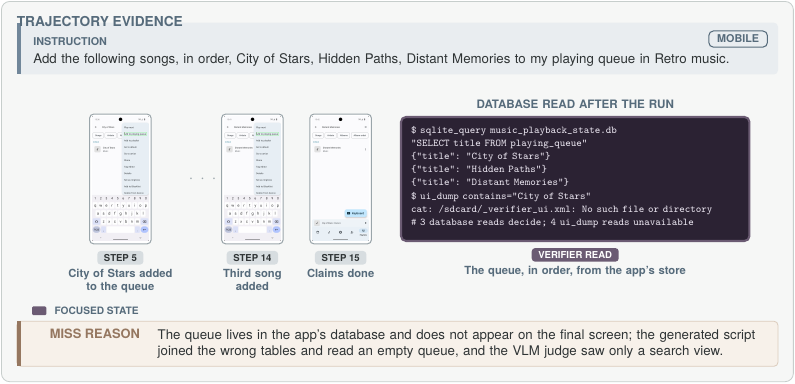}
{fig:case-retro}
{\textbf{Gold}\enspace\textsc{success}\quad\textbf{Agent}\enspace claimed complete\hfill
 \textbf{VLM judge}\enspace\textsc{fail}\quad\textbf{Generated verifier}\enspace\textsc{fail}\quad\textbf{VeriCUA}\enspace\textsc{success}}
{\rubricline{\rmet}{Each of the three songs is in the queue.}
 \rubricline{\rmet}{The songs follow each other in the asked order.}
 \rubricline{\rmet}{The queue holds exactly three songs.}}
{\evidenceline{\texttt{sqlite\_query} on the playback database, located
  through the environment map, persistent: the three titles in the asked
  order, on two tables.}
 \evidenceline{\texttt{ui\_dump}, volatile, four attempts: no dump was
  produced, so the verdict rests on the store alone.}}
{A success that only the app's store can attest. The queue is read from the
playback database in the asked order, while both baselines fail the run.}

\FloatBarrier
\Needspace{0.6\textheight}
\subsection{Recorded Trajectories}
\label{app:cases-recorded}

Here the verifier has nothing but the frames and logs the recording kept.

\casecard{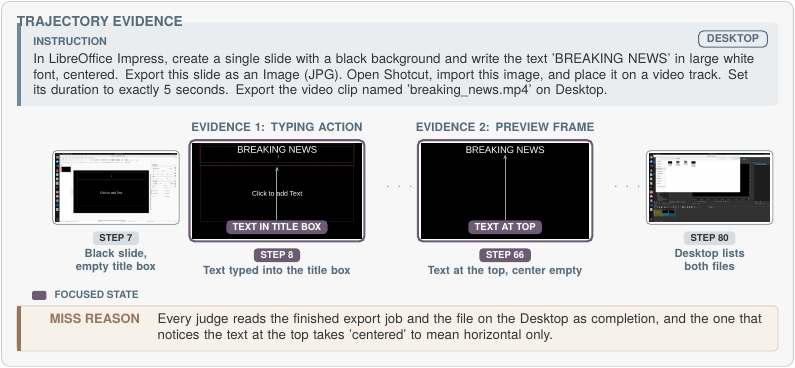}
{fig:case-shotcut}
{\textbf{Gold}\enspace\textsc{fail}\quad\textbf{Agent}\enspace claimed complete\hfill
 \textbf{TRM}\enspace\textsc{success}\quad\textbf{CUAJudge}\enspace\textsc{success}\quad\textbf{OS-Themis}\enspace\textsc{success}\quad\textbf{VeriCUA}\enspace\textsc{fail}}
{\rubricline{\runmet}{The text is centered in the video frame.}
 \rubricline{\runknown}{Six further requirements on the files, colors, and
  duration, left unsettled.}}
{\evidenceline{The preview frame at step 66, post-effect frame: the text
  sits at the top over an empty black field, read on all three paths.}
 \evidenceline{\texttt{inspect\_action} at step 7, post-effect frame: the
  text was typed into the slide's title placeholder, which fixed its
  position.}}
{A long desktop recording that every judge passes on the finished export.
The verifier reads the preview frame and finds the text at the top rather
than centered.}

\casecard{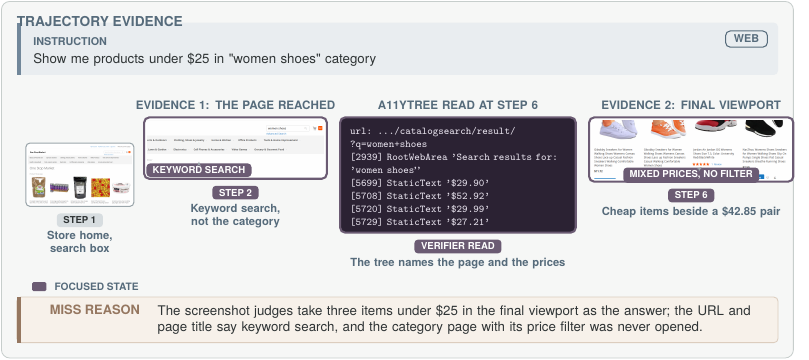}
{fig:case-shopping}
{\textbf{Gold}\enspace\textsc{fail}\quad\textbf{Agent}\enspace stopped without a claim\hfill
 \textbf{TRM}\enspace\textsc{success}\quad\textbf{CUAJudge}\enspace\textsc{success}\quad\textbf{OS-Themis}\enspace\textsc{success}\quad\textbf{VeriCUA}\enspace\textsc{fail}}
{\rubricline{\runmet}{The page is the women's shoes category.}
 \rubricline{\runmet}{Every listed price is under \$25.}}
{\evidenceline{\texttt{read\_a11y} at step 6, structured state: the URL and
  the page title name a keyword search for ``women shoes'', read on all three
  paths.}
 \evidenceline{The same tree lists several prices above \$25, so no price
  filter was applied.}}
{A web trajectory whose final screen shows cheap shoes. The \texttt{a11ytree} names the
page a keyword search rather than the category page and lists prices above the
limit.}

\casecard{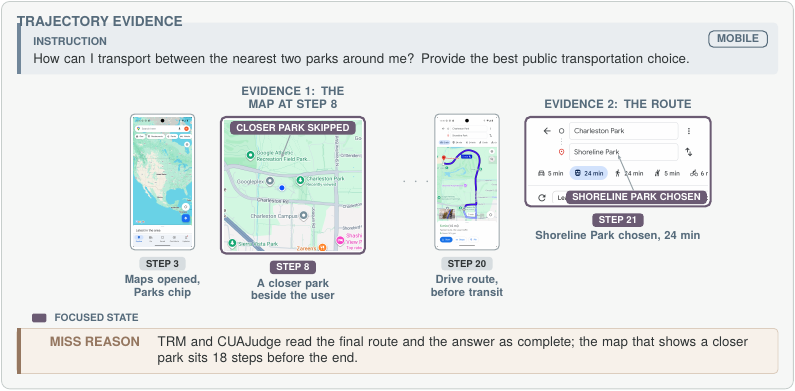}
{fig:case-maps}
{\textbf{Gold}\enspace\textsc{fail}\quad\textbf{Agent}\enspace answered, then terminated\hfill
 \textbf{TRM}\enspace\textsc{success}\quad\textbf{CUAJudge}\enspace\textsc{success}\quad\textbf{OS-Themis}\enspace\textsc{fail}\quad\textbf{VeriCUA}\enspace\textsc{fail}}
{\rubricline{\runmet}{The two parks nearest the user.}
 \rubricline{\runknown}{Both parks named, and the transit route or its
  absence stated, left unsettled.}}
{\evidenceline{The map at step 8, post-effect frame, fetched by every path:
  a park beside the user's location that the answer never mentions, while
  the chosen park lies 1.1 miles away.}
 \evidenceline{\texttt{zoom} at step 21, precise read: the route fields name
  Charleston Park and Shoreline Park.}}
{A mobile recording whose answer reads well. The map 18 steps before the end
shows a closer park than the one the agent chose, and two of three judges
passed it.}

\FloatBarrier
\endgroup

\FloatBarrier
\section{Limitations and Broader Impacts}
\label{app:lim}

\ours trades cost for evidence. A verification pass issues tens of tool calls
and model calls per trajectory, an order of magnitude more than a single-call
judge, so it suits filtering and reward assignment better than interactive
use. On recorded trajectories it abstains wherever the recording
holds no evidence for a requirement. Those abstentions are a coverage limit,
and online access lifts it only for environments that can be
instrumented. A stochastic verifier also needs its draw-to-draw spread reported beside every
comparison, as we do. Reliable trajectory verification lowers the cost of
training computer-using agents and of auditing benchmark labels. The same
machinery could be misused to certify agents that act on real systems, which
is one reason \ours never writes to the environment it verifies.
\FloatBarrier

\FloatBarrier
\Needspace{0.5\textheight}
\section{Prompts}
\label{app:prompts}

The system prompts below drive rubric synthesis in the two settings and
checker synthesis in online environments. Task-specific inputs such as
the instruction and the device context are appended at run time.

\FloatBarrier
\Needspace{0.35\textheight}
\subsection{Recorded-Trajectory Verification}
\label{app:prompts-recorded}

\begin{prompt}[H]
\centering
\begin{minipage}{0.96\textwidth}
\begin{Verbatim}[fontsize=\scriptsize, breaklines=true, breaksymbolleft={}, frame=single, framesep=4pt]
You decompose a computer-use task into atomic, individually checkable
requirements for a verifier. Output requirements about the FINAL END-STATE the
task demands (what a programmatic checker inspecting files/documents/settings
afterwards would verify), NOT about the method or the steps taken.

Rules:
- Each requirement is one concrete, verifiable condition.
- Honor every quantifier/detail: "all"/"each" => ALL; exact names/values/paths
  => exact. Split compound asks into separate requirements.
- Mark hard (must hold for success) or soft (minor/cosmetic).
- Do NOT invent process requirements ("must have clicked X"); if the end-state
  is right, the route does not matter.

Output ONLY:
```json
{"requirements": [{"id": "r1", "description": "...", "hard": true}, ...]}
```
\end{Verbatim}
\end{minipage}
\caption{Rubric synthesis on recorded trajectories.}
\label{prompt:rubric_recorded}
\end{prompt}

\FloatBarrier
\Needspace{0.55\textheight}
\subsection{Online Verification}
\label{app:prompts-online}

\begin{prompt}[H]
\centering
\begin{minipage}{0.96\textwidth}
\begin{Verbatim}[fontsize=\scriptsize, breaklines=true, breaksymbolleft={}, frame=single, framesep=4pt]
You author verification requirements for tasks executed on an Android device.
Given a task instruction, produce the END-STATE requirements a strict human
examiner would check AFTER the task supposedly finished.

Rules:
- END-STATE only: describe what must be true of the device state, never the
  procedure ("brightness setting equals maximum", not "user opened settings").
- Atomic: one checkable fact per requirement. Split name vs content vs location.
- Strict reading: if the instruction names a value, place, or format, it is
  required exactly; include implicit artifacts (a created note must exist AND
  contain the given text).
- hard=true for requirements whose failure fails the task; hard=false for
  cosmetic/soft ones.
- Only what the instruction implies: do NOT add adjacent settings, modes, or
  hygiene the instruction never mentions (no invented requirements).
- COUNTS ARE EXACT: "take one video" means exactly one new video was created,
  not at least one; "log the reimbursable transactions" means exactly that set
  was added — nothing missing, nothing extra. When the instruction quantifies,
  produce a requirement for the exact count or exact set of NEW artifacts
  (relative to the episode start), not mere existence.
- Do NOT bind platform-specific encodings or exact constants the instruction
  does not state: say "the minimum value the system allows", never "(0)".
  Resolving concrete encodings is the checkers' job, not yours.
- Do NOT bind storage locations or app conventions the instruction does not
  name ("a photo was taken" requires a new image in SOME standard media
  location, not specifically DCIM). Location is a requirement only when the
  instruction states one.
Output ONLY JSON:
```json
{"requirements": [{"id": "r1", "description": "...", "hard": true}]}
```
\end{Verbatim}
\end{minipage}
\caption{Rubric synthesis in online environments.}
\label{prompt:rubric_online}
\end{prompt}

\begin{prompt}[H]
\centering
\begin{minipage}{0.96\textwidth}
\begin{Verbatim}[fontsize=\scriptsize, breaklines=true, breaksymbolleft={}, frame=single, framesep=4pt]
You write CHECKERS that verify end-state requirements on an Android device.
A checker is DATA, not a script: a plan of read-only getter calls plus one
predicate over their outputs. You cannot write to the device; only these
getters exist:

{getter_docs}

{device_context}

Design rules:
- For each hard requirement produce 2 checkers probing DIFFERENT surfaces or
  routes where meaningful (e.g. settings vs dumpsys; direct path vs find).
  If only one sensible surface exists, 2 independent probes of it are fine.
- Prefer broad reads + predicate (list_dir or find + check the listing) over a
  single narrow existence probe: the task may have been satisfied at a
  different valid location or filename than the most literal guess.
- predicate kinds: "code" (one Python expression over the getter outputs)
  or "semantic" (a plain-language criterion decided by a separate judge).
  [rules on polarity, exact-name matching, store representations, and
  screen-only state omitted for space]
\end{Verbatim}
\end{minipage}
\caption{Checker synthesis in online environments, shortened where marked.}
\label{prompt:checker}
\end{prompt}
\FloatBarrier

\stopcontents[appendices]

\end{document}